\documentclass[manuscript, nonacm]{acmart}

\usepackage{amsmath}
\usepackage{xcolor}
\usepackage{soul}

\graphicspath{{./figures/}}

\DeclareMathOperator*{\argmin}{argmin}
\newcommand{\norm}[1]{\left\lVert#1\right\rVert}

\AtBeginDocument{%
  \providecommand\BibTeX{{%
    \normalfont B\kern-0.5em{\scshape i\kern-0.25em b}\kern-0.8em\TeX}}}

\setcopyright{acmcopyright}
\copyrightyear{2023}
\acmYear{2023}
\acmDOI{10.1145/1122445.1122456}

\acmConference[Woodstock '18]{Woodstock '18: ACM Symposium on Neural
  Gaze Detection}{June 03--05, 2018}{Woodstock, NY}

\begin{document}

\title{Text-to-Image Cross-Modal Generation: A Systematic Review}

\author{Maciej {\.Z}elaszczyk}
\email{m.zelaszczyk@mini.pw.edu.pl}
\orcid{0000-0002-9537-5985}
\author{Jacek Ma{\'n}dziuk}
\email{mandziuk@mini.pw.edu.pl}
\orcid{0000-0003-0947-028X}
\affiliation{%
  \institution{Warsaw University of Technology}
  \streetaddress{Plac Politechniki 1}
  \city{Warsaw}
  \country{Poland}
  \postcode{00-661}
}

\renewcommand{\shortauthors}{{\.Z}elaszczyk and Ma{\'n}dziuk}

\begin{abstract}
We review research on generating visual data from text from the angle of \textit{cross-modal generation}. This point of view allows us to draw parallels between various methods geared towards working on input text and producing visual output, without limiting the analysis to narrow sub-areas. It also results in the identification of common templates in the field, which are then compared and contrasted both within pools of similar methods and across lines of research. We provide a breakdown of text-to-image generation into various flavors of image-from-text methods, video-from-text methods, image editing, self-supervised and graph-based approaches. In this discussion, we focus on research papers published at 8 leading machine learning conferences in the years 2016-2022, also incorporating a number of relevant papers not matching the outlined search criteria. The conducted review suggests a significant increase in the number of papers published in the area and highlights research gaps and potential lines of investigation. To our knowledge, this is the first review to systematically look at text-to-image generation from the perspective of \textit{cross-modal generation}.
\end{abstract}

\maketitle

\section{Introduction}

In the wake of the 2012 ImageNet Large Scale Visual Recognition Challenge (ILSVRC), when AlexNet \cite{Krizhevsky2012} significantly outperformed all previous approaches, deep learning became the de facto standard for image classification, as it was able to achieve significantly higher accuracy levels than other methods. Convolutional neural networks (CNNs) have been the major force behind the advances in vision, first in their vanilla form with the use of backpropagation \cite{LeCun1998}, and then with the introduction of residual connections \cite{He2016}. This was mirrored by advances in natural language processing (NLP), which relied on recurrent neural networks (RNNs), with the Long Short-Term Memory (LSTM) \cite{Hochreiter1997} architecture becoming a particularly successful approach. Subsequently, the design of the attention mechanism \cite{Bahdanau2015} led to architectures based on attention, such as the Transformer \cite{Vaswani2017}, and the idea of pretraining for Transformers (BERT) \cite{Devlin2019}. The advances in Transformer architectures have entered the area of vision with the design of the Vision Transformer (ViT) \cite{Dosovitskiy2021}.

The CNN/ViT approaches for vision and the RNN approaches for NLP share a common thread of relying on classification tasks. This is more explicit for vision where the problem at hand is usually a classification task itself. NLP architectures may use classification in a more implicit way, for instance in machine translation where the problem formulation admits multi-step classification as a viable task formulation.

It has to be noted, however, that classification-based problems are not the only area of deep learning where significant progress has been made. One broad research field where new techniques have been successfully introduced is the area of generative models. The idea of using an encoder/decoder architecture \cite{Ackley1985} found renewed relevance for generative modeling with the arrival of variational autoencoders (VAEs) \cite{Kingma2014} and generative adversarial networks (GANs) \cite{Goodfellow2014}, followed by diffusion models \cite{Sohl-Dickstein2015}. The initially limited capabilities of such models were broadened, for instance by the introduction of training-stabilizing measures for GANs in the form of the DCGAN model \cite{Radford2016}. This was followed by more work extending the base approaches. Notable examples for VAEs would include the vector-quantized VAE (VQ-VAE) \cite{vandenOord2017} and its second iteration \cite{Razavi2019}. For GANs, improvements were made through models such as StackGAN \cite{Zhang2017stackgan} and its extensions \cite{Zhang2018}. With diffusion, progress was made with the advent of Denoising Diffusion Probabilistic Models (DDPMs) \cite{Ho2020}, \cite{Nichol2021}, \cite{Dhariwal2021}.

Zeroing in on VAEs, GANs and diffusion, all these approaches rely on random input as a seed for the data generation process. This randomness is the source of diversity in the generated samples. It has also been experimentally shown that the input might also carry a degree of semantic information about the generated samples. For vision, this could mean that the random data ingested into the system governs selected characteristics of what is visible in the produced images \cite{Radford2016}, and that those characteristics might correlate with the human understanding of the descriptive attributes.

This turns out to be true not only for random input, but also for specifically-tailored information provided to the architecture. In principle, VAEs, GANs and diffusion models do not strictly require all the input data to be random, so additional information can be shown to the model. In such a case, the generative process is conditioned on the input data and the architecture can be considered a conditional generative model. The precise form of the information to condition on may vary. In relatively simple settings, it could be the label of the class an instance of which we would want to generate. However, it need not be as simple.

\begin{figure}[h]
  \centering
  \includegraphics[width=0.8\linewidth]{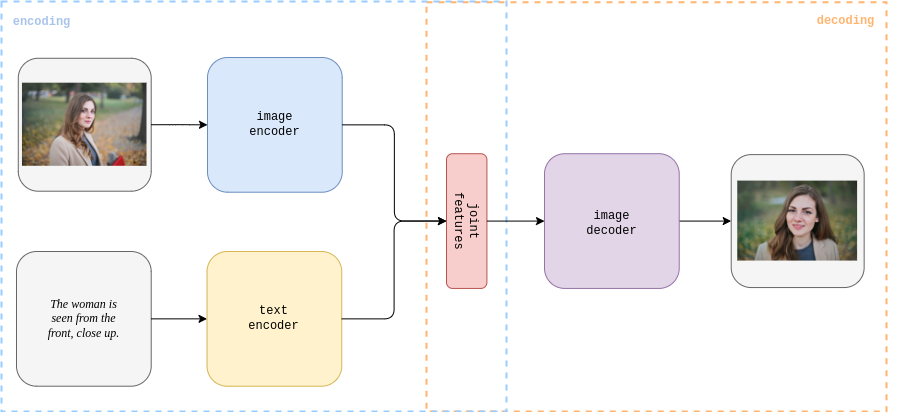}
  \caption{Conditioning image generation on additional data. An image is processed by an image encoder and the text describing the characteristics of the desired generated image is processed by a text encoder. The features obtained from both encoders are fused in a joint feature representation, which is passed along to the image decoder to produce the actual image.}
\label{fig:woman}
\end{figure}

If we consider the conditioning information to encode a subset of features of the desired generated image, then it is possible to use the actual features extracted either by separately-trained models or by upstream parts of a jointly-trained architecture. One relevant example could be the use of a CNN or ViT feature extractor (image encoder) to process an input image and arrive at the feature representation of this image. This representation could then be fed into a VAE-, GAN- or diffusion-based model. In principle, this process can be repeated for multiple data sources, with or without the inclusion of random data. A concrete example of this could entail an image editing task, where an input image is processed by a CNN/ViT image encoder and the text description of the desired final image is processed by an RNN text encoder. These representations are then fused into one and passed to a transposed convolutional (TCNN) \cite{Dumoulin2016} image decoder, a Transformer image decoder \cite{Vaswani2017}, potentially in an autoregressive manner \cite{vandenOord2016}, or to a diffusion image decoder \cite{Sohl-Dickstein2015} to produce the image which carries over the characteristics of the input image but corresponds to the provided description. The outline of such a process can be seen in Figure \ref{fig:woman}.

This example shows one important tenet of conditional data generation. Namely, there is no explicit assumption that the conditioning data comes from the same distribution, or in fact the same modality, as the output data. The description of the desired image comes from the text modality while the generated output is from the image modality. So, it is possible to generate data from one modality based on the input from another modality. This general process can be described as \textit{cross-modal generation}. Various possible cross-modal setups can be considered, with one or more input modality and, similarly, one or more output modality, where each of the input and output modalities could be different. It would then be possible, for instance, to generate audio from an image or, conversely, to generate an image from audio input.

Amongst the numerous possible modalities, the text and visual ones are the subject of significant research effort. This can be partially attributed to the fact that both the image and text modalities have been relatively intensely studied on their own with advances in vision and NLP, respectively. On top of that, the text or language domain possesses inherent structure due to the fact that most language problems rely on a limited vocabulary, which admits the use of mult-step classification methods for text processing. For example, generating a text description from an input image, known as \textit{image captioning}, might use a multi-step classification procedure for the selection of subsequent words in a caption. An important point here is that in this setup, both the image encoder and the text decoder can be used much as they are in vision and NLP, respectively, without significant modifications, which makes it easier to port those methods to the generative setting. Due to the above reasons, image captioning, which is a sub-field of image-to-text generation \cite{ZelaszczykMandziuk2023}, is potentially the most explored area within cross-modal text and visual generation. 

Going the other way, from the text domain to the visual domain (e.g. image, video, etc.) has received substantially less attention in terms of research output. An important reason behind this can be traced back to the inherent structure of the data. While text-to-image generation might reap benefits similar to image-to-text problems on the input end, where the structured nature of text can be exploited, the situation is entirely different on the output end, where images are concerned. Unlike text descriptions, images do not have a limited vocabulary, at least in the classic sense, and as such, the space of potential images is significantly larger than for text generation problems. This is due to the fact that for an image of a given size, the raw pixel output can be set at each point of the image, resulting in an explosive number of possible combinations, and a very high dimensionality of the problem. If we then consider the space of all \textit{valid} images, where by \textit{valid} we understand pictures which look like actual images from the dataset by some measure of similarity, randomly generating a valid image might be significantly less likely than randomly generating a valid sentence for the image captioning problem. By extension, while generating valid images using non-random methods is highly method-dependent, it might still be far more demanding than generating valid sentences based on images. All of this makes this problem far less amenable to multi-step classification methods. These difficulties have led to the research area of text-to-image generation being significantly underrepresented relative to image-to-text problems. Limited research output notwithstanding, significant progress has recently been made in this domain.

The image-to-text and text-to-image problems have been significantly expanded and are both at the forefront of research focusing on cross-modal generation. They have also seen the incorporation of research lines from other areas of deep learning. 

In this work, we specifically focus on the text-to-image problem and derivative tasks. With the increased interest and research output on the rise for this field, there is a need for a comprehensive review of the various strands of research.
To the best of our knowledge, the existing research on text-to-image generation lacks such a review and therein lies the main contribution of this work. We aim to establish links between various areas within text-to-image generation, as well as with other areas of deep learning, bringing together disjoint lines of research. It is our intention to unify the discussion from an overarching \textit{cross-modal generation} perspective.

The starting points of this review are the research papers published at 8 machine learning conferences:

\begin{itemize}
    \item Conference on Neural Information Processing Systems (NeurIPS)
    \item International Conference on Machine Learning (ICML)
    \item International Conference on Learning Representations (ICLR)
    \item AAAI Conference on Artificial Intelligence (AAAI)
    \item International Joint Conference on Artificial Intelligence (IJCAI)
    \item International Conference on Computer Vision (ICCV)
    \item European Conference on Computer Vision (ECCV)
    \item Conference on Computer Vision and Pattern Recognition (CVPR).
\end{itemize}

\begin{figure}[h]
  \centering
  \includegraphics[width=0.8\linewidth]{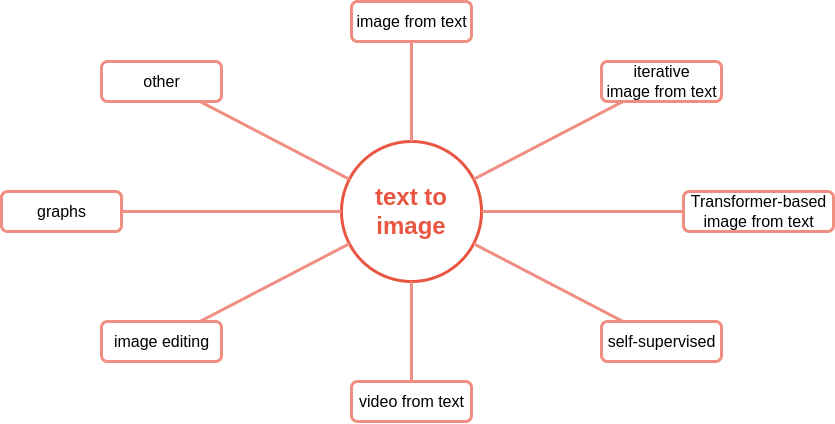}
  \caption{Cross-modal text-to-image generation research areas. These research areas are identified with respect to the architectures, training methods or tasks performed. The specific methods within each area can be significantly different, depending on concrete problems handled.}
\label{fig:topics-text-to-image}
\end{figure}

More concretely, we consider papers published in the time frame 2016-2022 - the most recent publication years at the moment of writing. We have searched the conference proceedings for the following terms: \textit{cross-modal}, \textit{multi-modal} (\textit{multimodal}), \textit{generative} and \textit{diffusion}. From papers conforming to these criteria, we select those that actually cover text-to-image generation. We also add a number of papers which do not meet the outlined search criteria but are still relevant as far as text-to-image generation is concerned, in particular, works on text-to-image diffusion models. We strive to provide a comprehensive review of cross-modal text-to-image generation, focusing on both the common elements of various methods, and their distinctive characteristics. The outline of the topics covered is presented in Figure \ref{fig:topics-text-to-image}. 

The flow of this review is structured as follows. In Section~\ref{sec:text-to-image}, the text-to-image generation problem is described, with a detailed discussion of the sub-areas related to this task. In particular, Section \ref{sec:image-from-text} covers the generation of images from text, Section \ref{sec:iterative-image-from-text} discusses iterative extensions to the standard approach, Section \ref{sec:transformer-image-from-text} focuses on Transformer-based variants, Section \ref{sec:self-supervised} describes self-supervised methods, Section \ref{sec:video-from-text} highlights the possibility of producing video from text input, Section \ref{sec:image-editing} tackles the task of editing an image based on a description, Section \ref{sec:graphs} considers graph methods, while Section \ref{sec:other} reviews the remaining idiosyncratic approaches. In Section \ref{sec:future}, potential lines of future research are discussed. Section \ref{sec:conclusions} concludes.

\section{Text to image}
\label{sec:text-to-image}

The overall process of generating images from text can be seen as related to the image captioning and visual dialogue tasks but it actually involves unique challenges. In simple terms, given one text description, a great multitude of images can be generated, all of them conforming to the provided description. This suggests that models which aim to produce images from text not only have to properly model the semantics of the text input and translate them into features useful for image generation, but also allow for significant diversity in the generated images, even in the case when the semantic features of the text do not change much, or at all. Compared with image captioning, generating images from text input can also be seen as more challenging on an additional level. For image captioning, it suffices to perform multiple-step classification over the vocabulary, where each step corresponds to one produced word or end-of-description sign. Going from text to images is more challenging in that it does not permit a simple classification task but rather requires the model to be able to generate images not only corresponding to the provided input but also retaining visual coherence and similarity to images seen in the wild. The requirement to convincingly model complex visual aspects has driven research efforts in this area to incorporate generative models such as GANs, VAEs or diffusion in the text-to-image architectures.

The generation of images or videos from text necessitates the use of a range of quantitative metrics for the assessment of the quality of the produced samples. Due to the ambiguity of the definition of quality as far as the produced output is concerned, specific metrics adopt varying approaches to the identification of quality and diversity. Broadly-used evaluation metrics include:

\begin{itemize}
    \item Inception Score (IS) \cite{Salimans2016}
    \item R-precision (top-1) \cite[Chapter~8.4]{Manning2008}
    \item $L_2$ error \cite{Nam2018}
    \item Fréchet Inception Distance (FID) \cite{Heusel2017}
        \begin{itemize}
            \item FID-$k$ is the Fréchet Inception Distance for images blurred by a Gaussian filter with radius $k$ \cite{Ramesh2021}, \cite{Ding2021}
        \end{itemize}
    \item Caption Loss (CapLoss) \cite{Ding2021}
\end{itemize}

A comparison of the evaluation results of the applicable described architectures on the Caltech-UCSD Birds dataset (CUB) \cite{Wah2011} and the MS COCO dataset \cite{Lin2014}, \cite{Chen2015} is shown in Table \ref{tab:cub-mscoco-generation}. It has to be noted that the evaluation metrics themselves are an active research area, with metrics such as Mutual Information Divergence (MID) \cite{Kim2022mutual} being proposed in order to improve the consistency across benchmarks, sample efficiency and the robustness of the evaluation to the choice of the models underlying the metric.

\subsection{Image from text}
\label{sec:image-from-text}

Generating images from text can be regarded as a reverse of the image captioning problem. A vanilla text-to-image generation task can be seen as one where text input is provided in the form of either a caption or a longer description and the goal of the model is to produce an image which would be assessed by human observers as corresponding to the input. In its simplest form, this task entails the processing of the text input by an RNN and using the obtained features to guide the part of the architecture responsible for producing the actual images.

The shape of the template for text-to-image methods depends on whether we consider a VAE-based method, a GAN-based method, a diffusion-based method, or perhaps a different kind of a model. One common component for all the templates is the RNN text encoder, which is used to process the provided description and obtain relevant semantic features. Another one would be the transposed convolutional neural network (TCNN) or a Transformer image decoder, which operates on features and produces an image based on those features.

\subsubsection{Variational autoencoder template}
\label{sec:vae-template}

In the VAE template, the features produced by the text encoder are used to obtain a representation, which is then processed by the image decoder. The system is trained based on a reconstruction loss with a term ensuring that the parameters used to sample random vectors adhere to a predetermined distribution.

This standard text-to-image VAE template translates from text into an image. An encoder is present, describing a distribution $q_{\phi}(\mathbf{f} \vert \mathbf{t})$, where $\mathbf{t}$ is the input text description used to obtain vector-valued parameters $\mu$ and $\sigma$, which are then used to sample from the latent space to obtain the \textit{visual-semantic features} $\mathbf{f}$. The sampling produces random representations $\mathbf{f} = \mu + \sigma \odot \epsilon$, where $\epsilon$ is sampled from a predetermined distribution, for instance $\epsilon \sim \mathcal{N}(\mathbf{0}, \mathbf{I})$, $\mathbf{I}$ is the identity matrix. The random representations are then processed by the decoder $p_{\theta}(\mathbf{i} \vert \mathbf{f})$, which maps to the image space. Training is based on the following loss:

\begin{equation}
    \mathcal{L}(\mathbf{t}, \mathbf{i}, \theta, \phi) = -\mathbb{E}_{\mathbf{f} \sim q_{\phi}(\mathbf{f} \vert \mathbf{t})}\left[\log{p_{\theta}(\mathbf{i} \vert \mathbf{f})}\right] + \mathbb{KL}(q_{\phi}(\mathbf{f} \vert \mathbf{t}) \vert\vert p(\mathbf{f}))
\end{equation}

\begin{figure}[h]
  \centering
  \includegraphics[width=\linewidth]{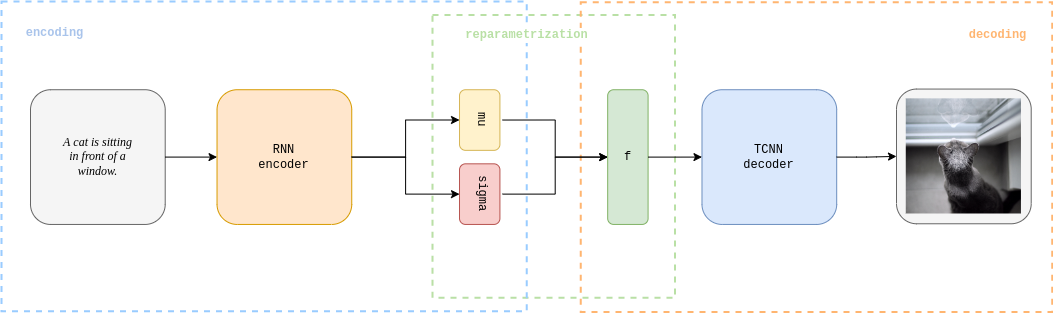}
  \caption{Standard image-from-text VAE template. The text describing the desired image is processed by an RNN encoder in order to obtain parameters $\mu$ and $\sigma$, which are used to sample the feature representation $\mathbf{f}$ of the image to be produced. This representation is used by a TCNN decoder to produce the actual image. A reconstruction loss is used in training to ensure the fidelity of the generated images to the dataset used for training, while a KL divergence term is used to enforce the alignment of $\mu$ and $\sigma$ with the parameters of a predetermined distribution.}
\label{fig:image-from-text-vae}
\end{figure}

Relative to the image captioning problem, the change here is in the ordering of the image $\mathbf{i}$ and text $\mathbf{t}$, which are now reversed. A schematic overview of the VAE template is presented in Figure \ref{fig:image-from-text-vae}.
\subsubsection{Generative adversarial network template}
\label{sec:gan-template}

For the GAN template, the image decoder is still used to produce an image from a representation obtained by means of the text encoder and random sampling. The image decoder is usually called the \textit{generator}. The difference is that a separate network, called the \textit{discriminator}, is used as an image processor to decide whether a given image belongs to the actual dataset or has been produced by the generator. The discriminator, apart from processing the image, also has access to the text features for a given real or generated image. The system is then trained adversarially where the generator and discriminator have opposing objectives.

A setup of modalities similar to the one for the VAE case is also seen for the GAN text-to-image template. The generator $G(\mathbf{f}) = G(\mathbf{z} \vert \mathbf{t})$ maps from the \textit{visual-semantic feature} space to the image space, $G: \mathbb{F} \to \mathbb{I}$, where $\mathbf{f} = \left( \mathbf{z}, \mathbf{t} \right) \in \mathbb{F}$. The \textit{visual-semantic features} include the text features $\mathbf{t}$ and a random component from the latent space $\mathbf{z}$. $\mathbb{F}$ is the \textit{visual-semantic feature} space, $\mathbb{I}$ is the visual space of possible images. The sampling of $\mathbf{z}$ is done from a pre-determined distribution, e.g. $\mathbf{z} \sim \mathcal{N}(\mathbf{0}, \mathbf{I})$, where $\mathbf{I}$ is the identity matrix. The discriminator $D(\mathbf{i} \vert \mathbf{t})$ tries to assess whether a sample is real or produced by $G$. The text features $\mathbf{t}$ are supplied to $D$, $D: \mathbb{U} \to (0, 1)$, where $\mathbf{u} = \left[ \mathbf{i}, \mathbf{t} \right]^{T} \in \mathbb{U}$. $\mathbb{U}$ is the joint space of image and text representations. Training proceeds as a two-player minimax game, where the generator and discriminator push the objective function into opposite directions:

\begin{equation}
    \min_{G}\max_{D}V(D,G) = \mathbb{E}_{\mathbf{i} \sim p_{data}(\mathbf{i})}\{\log{(D(\mathbf{i} \vert \mathbf{t}))}\}
    +\mathbb{E}_{\mathbf{z} \sim p_{\mathbf{z}}(\mathbf{z})}\left[\log{(1-D(G(\mathbf{z} \vert \mathbf{t})))}\right]\}
\end{equation}

\begin{figure}[h]
  \centering
  \includegraphics[width=0.8\linewidth]{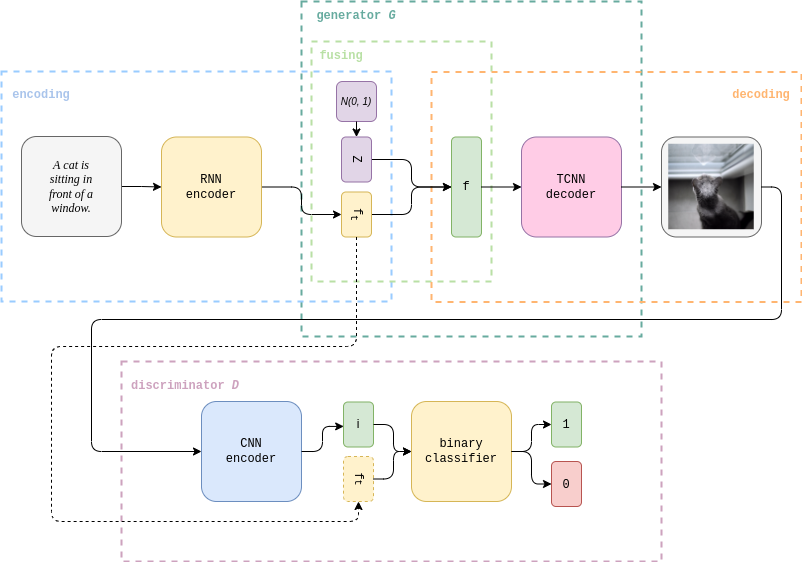}
  \caption{Standard image-from-text GAN template. The text description of the desired image is processed by an RNN encoder to obtain the features of the text $\mathbf{f_{t}}$. A random component $\mathbf{z}$ is sampled separately from a predetermined distribution. In the generator part of the architecture, both $\mathbf{f_{i}}$ and $\mathbf{z}$ are combined together in a joint representation $\mathbf{f}$, which is fed to a TCNN decoder to produce the image. The discriminator part of the architecture handles such produced images, as well as real images from the dataset. An image is processed by a CNN encoder to arrive at the image features $\mathbf{i}$. These image features, together with the text features $\mathbf{f_{t}}$ are fed into a binary classifier which aims to distinguish whether the processed image is real (comes from the dataset) or fake (has been produced by the generator). During training, the generator and discriminator compete to push the objective function into different directions.}
\label{fig:image-from-text-gan}
\end{figure}

An outline of the GAN template is shown in Figure \ref{fig:image-from-text-gan}.

\subsubsection{Diffusion template}
\label{sec:diffusion-template}

The diffusion template employs an RNN text encoder to process the semantic information in a prompt - a similarity to the VAE and GAN templates. While initial diffusion models operated directly in the pixel space of the image, it has been shown that applying diffusion to the latent space may yield advantages in terms of both sample quality and computational efficiency \cite{Rombach2022}. Because of that, we will focus on \textit{latent diffusion models} in the formulation of the diffusion template. In this form, the diffusion template makes use of an image encoder and an image decoder. The image representation obtained from the encoder is subjected to a forward diffusion process for a number of steps. This diffusion process is then reversed through a denoising procedure, where each step is conditioned on the text features. A reconstruction loss guides the training procedure.

The image encoder models the distribution $q_{\phi}(\mathbf{f} \vert \mathbf{x})$ where we obtain the \textit{visual-semantic features} $\mathbf{f}$ from the image $\mathbf{x}$, albeit without explicit conditioning on the text features $\mathbf{t}$. An image decoder $p_{\theta}(\mathbf{x} \vert \mathbf{f})$ is used to produce an image from the \textit{visual-semantic features}, still without explicitly conditioning on the text features. The encoder and decoder can be thought of as an encoder-decoder trained in the first stage to ensure its capabilities to reconstruct images. In the second stage, the obtained \textit{visual-semantic features} are used as an input to a forward diffusion process where representations $\mathbf{f}_{1}, \dots, \mathbf{f}_{T}$ are obtained by progressively adding noise to the initial features. In the reverse process, the aim is to sequentially recover the initial representation by applying a denoising autoencoder $\epsilon_{\rho}(\mathbf{f}_{t}, t, \mathbf{t})$, where $t = 1, \dots, T$. Importantly, each step of this procedure is conditioned on the text features $\mathbf{t}$. The second-stage training is performed using the following loss formulation:

\begin{equation}
    \mathcal{L}(\mathbf{t}, \mathbf{x}, \rho) = -\mathbb{E}_{p(\mathbf{x}), \mathbf{t}, \epsilon \sim \mathcal{N}(0, 1), t} \left[ \norm{\epsilon - \epsilon_{\rho}(\mathbf{f}_{t}, t, \mathbf{t})}^{2}_{2}\right]
\end{equation}

where $\epsilon$ is the sampled noise.

\begin{figure}[h]
  \centering
  \includegraphics[width=0.85\linewidth]{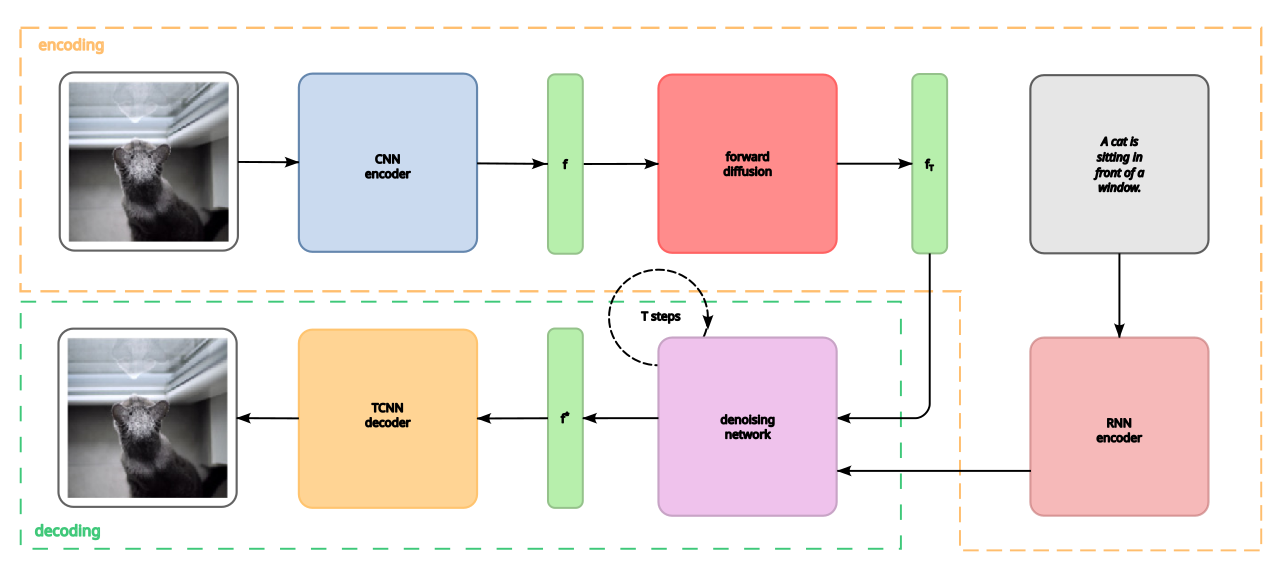}
  \caption{Standard image-from-text diffusion template. Training: An image is processed by an image encoder to obtain features $\mathbf{f}$. These features are subjected to a forward diffusion process where noise is gradually added to them, resulting in the final noise $\mathbf{f_{T}}$, which is passed to the denoising network along with the conditioning information from the RNN encoder. The denoising network progressively removes the noise from the representation to finally arrive at the reconstructed representation $\mathbf{f^{*}}$, which is processed by the image decoder to produce the reconstructed image. The optimization procedure attempts to successfully reconstruct the input image. Inference: instead of $\mathbf{f_{T}}$ obtained through forward diffusion, randomly sampled noise is used as input to the denoising network, along with the output of the RNN text encoder.}
\label{fig:image-from-text-diffusion}
\end{figure}

The recovered \textit{visual-semantic features} $\mathbf{f}$ are used as input to the image decoder to generate sample images. A sketch of the diffusion template is presented in Figure \ref{fig:image-from-text-diffusion}.

\subsubsection{Specific examples}
\label{sec:specific-examples}

Based on the described three templates, it is visible that the individual proposed approaches for text-to-image generation might differ substantially. Also, the VAE, GAN and diffusion templates are not the only approaches possible. Other generative models can be used for this task, as well as non-generative ones.

In one of the first works successfully applying adversarial learning to the text-to-image generation problem, \cite{Reed2016synthesis} show that it is possible to train a GAN model to generate images from text. Their architecture fits in the GAN template, with the text encoding supplied to both the generator and the discriminator. One difference is that rather than a standard RNN text encoder, they propose a CNN-RNN text encoder, which is pre-trained specifically to obtain text feature representations which can also be used for visual tasks. Additionally, they discuss extensions based on text/image matching, text manifold interpolation and an inverse network for style transfer. Three datasets are used for experiments: Oxford-102 \cite{Nilsback2008}, CUB \cite{Wah2011} and MS COCO. Depending on the dataset, the experimental results vary in which systems are able to generate realistic images matching the descriptions. For the Oxford-102 dataset, all the analyzed methods succeed. For the CUB dataset, there is more ambiguity. In this case, a base system trained without any enhancements struggles to produce real-looking images. The same goes for a system with text/image matching alone (i.e. without other enhancements). On the other hand, systems with interpolation or with both interpolation and text/image matching are able to generate relatively plausible images partially matching the captions. For the MS COCO dataset, the results from the text/image matching variant look superficially appealing, but the generated images lack in scene coherence. The obtained results highlight the challenges related to image generation and the multiple requirements the produced images have to satisfy. 

While GANs, VAEs and diffusion are some of the most popular choices as far as text-to-image generation is concerned, it is possible to construct models not relying on those methods. \cite{Mansimov2016} extend the DRAW architecture to generate images from captions. This architecture is substantially different than the VAE or GAN templates. They use a bidirectional RNN for caption encoding, which is the text encoder in this case, and an RNN which iteratively modifies the output image while attending to the text representation at each step when the image being generated is modified. After the final generative step, the canvas matrix is additionally sharpened by a separate network to produce the final image. The iterative RNN together with the sharpening network can be thought of as the image decoder. Experiments on the MS COCO dataset show that the architecture is able to generate images conditioned on the provided text and that it retains some features from the text representation, such as colors or background type. At the same time, the model struggled to handle situations where the described objects were similar, generating generic shapes. The system, however, displayed an ability to generate images from captions describing situations not found in the dataset, or even in real life.

Similarly, \cite{Tan2019} explicitly design an architecture without the GAN component. Based on RNNs, CNNs, CNN-RNNs and attention, this architecture is able to generate images and other kinds of data (such as object layouts). Since it is not based on VAEs or GANs, the architecture is considerably different than the standard VAE or GAN templates. It does incorporate an RNN text encoder to produce text representations, which is the only part from the standard templates that is directly utilized. Apart from that, it relies on iterative image modification. A CNN image encoder is used to represent the features of the state of the generated image. CNN-RNN modules with attention process the information from the image and text encoders to predict objects and their attributes. An optional component generates a representation which is compared with representations obtained from actual images in the dataset. The model handles producing abstract scenes for the Abstract Scenes dataset \cite{Zitnick2013}, as well as semantic layout prediction and composite image generation for MS COCO. For the image generation task, the experimental results point to the fact that the approach is competitive relative to other methods, e.g. AttnGAN \cite{Xu2018}, while at the same time not being a GAN method.

As mentioned before, VAEs are a popular choice for cross-modal generation and the text-to-image domain is no different. \cite{Yan2016} propose a VAE-based generative architecture for generating images from text. This is done via a procedure in which an image is paired with its attribute description and then the attributes are used to generate the image. One siginificant change relative to the standard VAE approach is that an RNN encoder is not used at all. Rather than apply such an encoder to the description of an image, a description in terms of a prediction of the attributes of the image is taken as given. In practice, this means using a vector with attribute predictions obtained from an external source, for instance from an attribute prediction system, which operates on the base of multilabel classification. This attribute representation is then processed by the decoder part of the system. An additional deviation from the standard VAE method is the fact that the image is modeled as a composite of foreground and background through a layered conditional VAE. This means that rather than a single CNN image encoder and TCNN image decoder, two separate CNN encoders are used for foreground and background encoding, and two separate TCNN decoders are used for foreground and background decoding, respectively. The overall architecture generates images from attributes rather than simple text descriptions, which itself is a significant modification. Also, it has the capacity to disentangle the generation of the foreground from the generation of the background. Experimental results on the LFW \cite{Huang2008} and CUB datasets demonstrate that the architecture is able to generate relatively high quality and diverse images. The results suggest that visual attributes and the disentangling of the foreground and background help in achieving higher quantitative scores.

Attempts to enrich the data structures used by the generative models and improve the quality of the produced samples are an important part of the research landscape in text-to-image generation. The text-to-image translation problem is approached by \cite{Ma2018} from the point of view of both set-level and instance-level characteristics. Their GAN-based architecture is designed to decompose the task into translating instances in a latent space with additional structure. In this way, both instance- and set-level characteristics are used in the generative process. While the overall architecture is GAN-based, it veers significantly off the standard GAN template path and actually incorporates elements from the VAE template. As far as the similarities to the GAN template are concerned, the proposed system still uses a CNN encoder, albeit with modifications, and it does utilize a TCNN decoder to produce the image. The system is also trained adversarially with the use of a discriminator. The differences relative to the standard approach would include the fact that the image encoder is attention-based rather than a vanilla CNN, inputs for the source and the target are processed together, and instance-level correspondences are obtained for the source and target input data instead of a vanilla visual feature representation for a single input. Reconstruction losses for both the translation from source to target and from target to source domains are included, taking inspiration from the standard VAE approach. These losses are used in the adversarial training procedure where two discriminator networks are employed to enforce set-level consistency. The experimental results on a variety of tasks not limited to text-to-image generation show that the proposed method performs favorably to a set of baselines. In particular, the architecture is able to outperform a very strong baseline, i.e. StackGAN \cite{Zhang2017stackgan}, in quantitative evaluations for the text-to-image generation task on the CUB dataset.

Diffusion models are employed by \cite{Esser2021imagebart} in order to perform a variety of image generation tasks, in particular, text-to-image generation. The general outline of the approach adheres to the standard diffusion template. A CNN encoder is used to obtain a representation of the image and a CNN decoder is used to generate samples. Those two networks are trained together in a VQGAN \cite{Esser2021taming} setup. The text representation is obtained via the CLIP model \cite{Radford2021}. The main difference relative to the standard template is the use of a multinomial forward diffusion process and the fact that the reverse diffusion process is modeled specifically via an autoregressive Transformer encoder-decoder architecture. Results on the Conceptual Captions dataset \cite{Sharma2018} indicate that the proposed model is able to improve upon VQGAN \cite{Esser2021taming} in terms of quantitative evaluation metrics.

A prominent line of work within cross-modal generative models focuses on the ability of VAEs to form joint representations of different modalities \cite{Wu2018mvae}, \cite{Shi2019}. For text-to-image, these approaches can be seen as incarnations of the standard VAE-based template. \cite{Joy2022} provide an extension of this line of work, with a VAE-based model capable of text-to-image generation. Their setup avoids the assumption that the representation from one modality can be wholly summarized by the representations from another one. Instead, the architecture relies on a generative model and a recognition model, which are based on encoders and decoders. The generative and recognition models are used in two modes. In one of them, the information flows from one modality to the other, and in the other mode, the flow is reversed. Similarly, the parameters of the generative and recognition models are swapped between the regular and mirrored flow. Experiments on the CUB dataset show that the proposed method is able to obtain higher quantitative evaluation scores than the baselines of MVAE \cite{Wu2018mvae} and MMVAE \cite{Shi2019}.

Unlike most of the published research on text-to-image generation, \cite{Daunhawer2022} start from the point of view of identifying the underlying limitations of existing architectures rather than proposing one. More specifically, they focus on the cross-modal VAEs and, by extension, on the standard text-to-image VAE template. A proof is provided that the sub-sampling of modalities introduces an irreducible discrepancy into the optimization process. Results on the CUB dataset point to the fact that VAE-based methods with modality sub-sampling (e.g. MMVAE \cite{Shi2019} or MoPoE-VAE \cite{Sutter2021}) display a quality gap relative to unimodal VAEs as well as cross-modal VAE-based methods without modality sub-sampling (e.g. MVAE \cite{Wu2018mvae}). This is particularly pronounced for larger values of the regularization constant $\beta$, which regulates the importance of the KL divergence term in the optimization procedure. Additionally, the results indicate that for datasets such as CUB, the generative coherence of MVAE increases markedly for high values of $\beta$, which suggests further potential for the investigation of the properties of methods without modality sub-sampling.

The demarcation line between the VAE and GAN templates is not a hard one, particularly as VAE-based models might be learned adversarially, much like GANs. \cite{Chen2022} adopt an approach where their architecture has the core elements of the vanilla VAE method, however, it does also include GAN-based components, such as discriminators and the adversarial learning procedure. The framework extends the line of research focused on encoder-decoder methods and probability factorization \cite{Wu2018mvae}, \cite{Shi2019}, \cite{Sutter2020}, \cite{Sutter2021}. The main distinguishing features are the alignment of multiple separate encoder distributions with a joint decoder distribution and the use of a specific factorization of the discriminator, which is amenable to contrastive training. Canonical Correlation Analysis (CCA) results on the CUB dataset suggest that the described model may perform favorably to MVAE \cite{Wu2018mvae} and MMVAE \cite{Shi2019}. 

\cite{Huang2022} approach the task of text-to-image generation assuming that more than one modality can be present on the input end. Their general architecture follows the standard GAN template with one significant difference being the ability to handle multiple input modalities. This leads to the model handling multiple input encoders in a \textit{product-of-experts} \cite{Hinton2002} fashion. Additionally, the ability to handle latent spaces with varying resolutions is added as a hierarchical component. One generator and one discriminator are still used, as in the vanilla text-to-image template. Both, however, are significantly modified. The generator consists of residual blocks, where latent features are sampled locally. The discriminator receives an image and a set of information about the input modalities rather than the image and the input text features as is the case in the standard GAN template. It is also adapted to handle multiple scales in the latent space. Quantitative results of the evaluation on MS COCO and the Multi-Modal CelebA-HQ dataset show that the proposed method significantly outperforms a range of baselines, including VQGAN \cite{Esser2021taming} on MS COCO and TediGAN \cite{Xia2021} on the Multi-Modal CelebA-HQ dataset.

A slightly different take on diffusion text-to-image generation is presented by \cite{Blattmann2022}. The main point is to use a semi-parametric generative model, e.g. a diffusion one, which is small relative to standard diffusion or autoregressive approaches. This model is trained via conditioning on sets of images from an external database, where these sets are selected such that they comprise nearest neighbors of the input image in the embedding space. During inference, the conditioning database can be swapped out for a different one. Using a CLIP encoder \cite{Radford2021} allows for conditioning not only on nearest neighbor images but also on text representations and a mixture of the two. In experiments, the model trained on the dogs subset of ImageNet \cite{Deng2009} with 20 million cropped images from OpenImages \cite{Kuznetsova2020} performs best on unconditional generation. Conditional text-to-image capabilities are evaluated on the MS COCO dataset and compared with LAFITE \cite{Zhou2022}. Quantitative evaluations show that the proposed method performs favorably to LAFITE. A qualitative assessment of generated samples suggests that conditioning on text representations alone results in better generalization capabilities to unseen prompts than conditioning on nearest neighbor images or on both nearest neighbor images and text.

Improving the diffusion model itself is not the only way to tackle text-to-image synthesis. \cite{Saharia2022} show that it may be crucially important to focus on the text encoder itself. Their general architecture follows the diffusion template, with the important distinction being that they do not use a latent diffusion model but rather a regular one in pixel space. Their focus on the language model leads to the experiment conclusion that using a generic large language model, in particular the T5 \cite{Raffel2020} Transformer architecture, can provide significant gains in terms of quantitative metrics on the MS COCO dataset relative to DALL-E \cite{Ramesh2021}, LAFITE \cite{Zhou2022}, GLIDE \cite{Nichol2022}, DALL-E 2 \cite{Ramesh2022}. Additionally, a new set of prompts for evaluating the performance of text-to-image models is introduced under the name DrawBench. Human evaluators on the DrawBench benchmark show a strong preference for the proposed method relative to the mentioned alternative models. There also seems to be a strong preference for T5 encoders relative to CLIP \cite{Radford2021} ones on this benchmark. These results support the claim that scaling up the text encoder is perhaps more important than scaling up the diffusion model itself.

The idea of composing diffusion models is explored in \cite{Liu2022compositional}. The backbone of the method adheres to the standard diffusion template, however, diffusion in pixel space rather than latent space is used. By analogy to energy-based models, diffusion models are combined. In practice, this boils down to a standard diffusion procedure with the major deviation being that an image in the reverse diffusion process is processed in parallel with different text context, where each context represents an element of the composition. The representations obtained for each context are then combined and produce the image at a given time step. The combination of different elements takes place based on two defined operators: AND and NOT. Three datasets are used for experiments: CLEVR \cite{Johnson2017}, Relational CLEVR \cite{Liu2021}, FFHQ \cite{Karras2019}. The designed method is compared against a set of baselines: EBM \cite{Du2020}, StyleGAN 2 \cite{Karras2020}, LACE \cite{Nie2021} and GLIDE \cite{Nichol2022}. The method shows stronger quantitative results than all the considered baselines on the CLEVR dataset. Additionally, a GLIDE-based compositional model generates samples which are quantitatively significantly more consistent with the provided prompts than the base GLIDE model.

While initial diffusion models for images \cite{Sohl-Dickstein2015} and their subsequent improvements in the form of Denoising Diffusion Probabilistic Models (DDPMs) \cite{Ho2020}, \cite{Nichol2021}, \cite{Dhariwal2021} operate in pixel space, \cite{Rombach2022} explore the possibility of operating in the latent space and initiate the research line of \textit{latent diffusion} models for vision. This approach is largely what the standard diffusion template is modeled after. The proposed technique achieves competitive performance on a range of tasks. In particular, when conditioning on LAION-400M \cite{Schuhmann2021} text prompts, the model quantitatively outperforms DALL-E \cite{Ramesh2021}, CogView \cite{Ding2021} and LAFITE \cite{Zhou2022} in evaluations on the MS COCO dataset. It is also found that the application of classifier-free guidance \cite{Ho2021} significantly improves performance.

Diffusion models can be chained with other types of architectures to form specific pipelines. \cite{Gu2022} use a VQ-VAE-based \cite{vandenOord2017} image encoder/decoder and model its latent space with diffusion. In this respect, the method is different from the standard diffusion template by utilizing quantized representations but similar in the fact that it is the latent representation that is modeled. More concretely, VQGAN \cite{Esser2021taming} is used as the specific  image encoder/decoder model and CLIP \cite{Radford2021} is used as the text encoder. Such an approach allows the model to significantly limit the number of trainable parameters relative to autoregressive setups. On the MS COCO, CUB and Oxford-102 datasets, the model is compared against a range of strong baselines, including DALL-E \cite{Ramesh2021} and CogView \cite{Ding2021}. Quantitatively, the model achieves performance superior to all the considered baselines and all three datasets where evaluation results of other methods are available. Qualitatively, samples from this model are compared to DM-GAN \cite{Zhu2019} and DF-GAN \cite{Tao2022}, and shown to be of favorable quality.

While Transformer-based autoregressive models have shown significant text-to-image generation capabilities, diffusion approaches have challenged this status. \cite{Ramesh2022} build upon previous work to use a CLIP \cite{Radford2021} text and image encoder and use a pre-trained CLIP model together with a newly trained image decoder. The proposed approach sits at a boundary line between autoregressive and diffusion-based methods. It utilizes a diffusion model, specifically a modified version of GLIDE \cite{Nichol2022} as the image decoder. At the same time, it employs one of two potential models for the representation passed to the image decoder: an autoregressive prior or a latent diffusion model. The unreleased datasets from the CLIP \cite{Radford2021} and DALL-E \cite{Ramesh2021} models are used in the training procedure. Various configurations of the architecture are evaluated. Quantitative results for text-to-image generation on the MS COCO dataset suggest that the diffusion prior is slightly more suitable than the autoregressive one and both versions beat the baselines of DALL-E \cite{Ramesh2021}, LAFITE \cite{Zhou2022} and GLIDE \cite{Nichol2022} in a zero-shot setting. Human evaluations show that users prefer the diffusion model to the autoregressive one and the proposed approach is scored significantly higher than GLIDE \cite{Nichol2022} in terms of diversity, while yielding comparable scores in terms of photorealism and the similarity of the image to the provided prompt.

\begin{table}
  \caption{Text-to-image generation, CUB \& MS COCO. The presented evaluation metrics are described in Section \ref{sec:text-to-image}.}
  \label{tab:cub-mscoco-generation}
  \resizebox{\textwidth}{!}{
    \begin{tabular}{ccccccccccc}
      \toprule
      \multicolumn{7}{c}{CUB \cite{Wah2011}} \\
      \midrule
      method & IS \cite{Salimans2016} & R-precision (top-1) \cite[Ch.~8.4]{Manning2008} & $L_2$ error \cite{Nam2018} & FID \cite{Heusel2017} & iterative & comments \\
      \midrule
      \cite{Zhang2017stackgan} & 3.70 $\pm$ 0.04 & - & - & - & \checkmark & - \\
      \cite{Bodla2018} & 3.92 $\pm$ 0.05 & - & - & - & \checkmark & - \\
      \cite{Xu2018} & 4.36 $\pm$ 0.03 & 67.82 $\pm$ 4.43 & - & - & \checkmark & - \\
      \cite{Li2019controllable} & 4.58 $\pm$ 0.09 & 69.33 $\pm$ 3.23 & 0.18 & - & \checkmark & - \\
      \cite{Zhu2019} & 4.75 $\pm$ 0.07 & 72.31 $\pm$ 0.91 & - & 16.09 & \checkmark & - \\
      \cite{Qiao2019} & 4.56 $\pm$ 0.05 & 57.67 & - & - & \checkmark & - \\
      \cite{Yin2019} & 4.67 $\pm$ 0.09 & - & - & - & \checkmark & - \\
      \cite{Cheng2020} & 5.23 $\pm$ 0.09 & - & - & - & \checkmark & - \\
      \cite{Gu2022} & - & - & - & 10.32 & - & - \\
      \midrule
      \multicolumn{7}{c}{MS COCO \cite{Lin2014}, \cite{Chen2015}} \\
      \midrule
      method & IS \cite{Salimans2016} & R-precision (top-1) \cite[Ch.~8.4]{Manning2008} & $L_2$ error \cite{Nam2018} & FID \cite{Heusel2017} & iterative & comments \\
      \midrule
      \cite{Tan2019} & 24.77 $\pm$ 1.59 & - & - & - & - & - \\
      \cite{Zhang2017stackgan} & 8.45 $\pm$ 0.03 & - & - & - & \checkmark & - \\
      \cite{Xu2018} & 25.89 $\pm$ 0.47 & 85.47 $\pm$ 3.69 & - & - & \checkmark & - \\
      \cite{Li2019controllable} & 24.06 $\pm$ 0.60 & 82.43 $\pm$ 2.43 & 0.17 & - & \checkmark & - \\
      \cite{Zhu2019} & 30.49 $\pm$ 0.57 & 88.56 $\pm$ 0.28 & - & 32.64 & \checkmark & - \\
      \cite{Qiao2019} & 26.47 $\pm$ 0.41 & 74.52 & - & - & \checkmark & - \\
      \cite{Yin2019} & 35.69 $\pm$ 0.50 & - & - & - & \checkmark & - \\
      \cite{Liang2020} & 52.73 $\pm$ 0.61 & 93.59 & - & - & \checkmark & - \\
      \cite{Zhang2021contrastive} & 30.45 & 71.00 & - & 9.33 & - & self-supervised \\
      \cite{Ramesh2021} & 17.90 & - & - & 17.89 & - & zero-shot, as reported by \cite{Ding2021}, \cite{Saharia2022} \\
      \cite{Ding2021} & 18.20 & - & - & 27.1 & - & zero-shot \\
      \cite{Huang2022} & - & - & - & 20.5 & - & - \\
      \cite{Blattmann2022} & 24.31 & - & - & 22.08 & - & zero-shot \\
      \cite{Nichol2022} & - & - & - & 12.24 & \checkmark & zero-shot \\
      \cite{Saharia2022} & - & - & - & 7.27 & - & zero-shot \\
      \cite{Saharia2022} & 26.62 $\pm$ 0.38 & - & - & 12.61 & - & zero-shot \\
      \cite{Gu2022} & - & - & - & 13.86 & - & - \\
      \cite{Ramesh2022} & - & - & - & 10.39 & - & zero-shot \\
      \bottomrule
    \end{tabular}
  }
\end{table}

\subsection{Iterative image from text}
\label{sec:iterative-image-from-text}

A specific line of research in text-to-image generation employs the idea of sequentially generating feature representations or images themselves and refining those representations or images. Such architectures may employ pipelines in which images of progressively higher resolution are generated or ones in which relatively crude representations are produced first before being expanded into richer ones. The sequential nature of the process might also be expressed by hierarchical components of the architecture. Examples include systems based on architectures such as VQ-VAE \cite{vandenOord2017}, StackGAN \cite{Zhang2017stackgan} or a number of their extensions.

The basic outline of an iterative method is based on the standard template for vanilla image-from-text methods, be it VAE- or GAN-based, and it shares the basic components with it. We do have an RNN text encoder, the information from which is fused with the randomly generated samples from the latent space. In the VAE variant, we have a TCNN image decoder, and the whole system is trained using a VAE objective. The difference here is that instead of one TCNN image decoder, we do have several ones, and the output of the prior ones is used by ones further down the pipeline. For the GAN variant, apart from the RNN encoder, we have a TCNN image generator and a CNN image discriminator. Again, the difference lies in the fact that we have more than one generator and more than one discriminator, and that the images produced by the generators at the latter stages of the pipeline utilize information from the images generated earlier on in the pipeline. For both the VAE and GAN variants, the precise setup of image decoders/generators and discriminators will vary with the specific method.

\begin{figure}[h]
  \centering
  \includegraphics[width=0.85\linewidth]{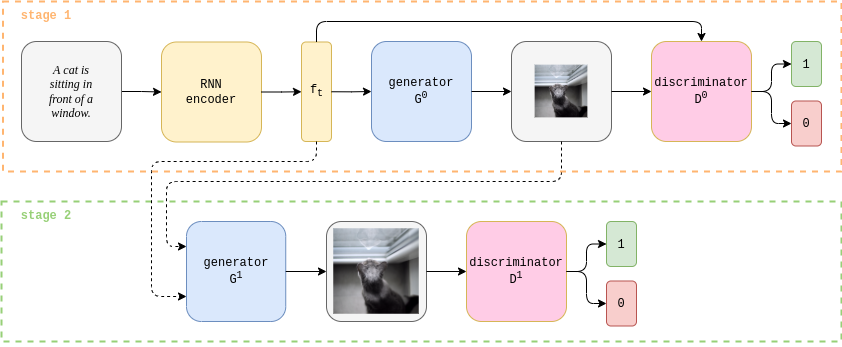}
  \caption{Standard iterative image from text GAN template. An input text description is processed by an RNN encoder to obtain the feature representation $\mathbf{f_{t}}$ of the text. The first-stage generator $\mathbf{G^{0}}$ uses this representation to produce an image. This image, along with the text features, is fed into the first-stage discriminator $\mathbf{D^{0}}$, which is tasked with distinguishing real images from the ones produced by $\mathbf{G^{0}}$. Both the first-stage image and the text features are used by the second-stage generator $\mathbf{G^{1}}$ to produce a refined version of the image. This refined image is then assessed by the second-stage discriminator $\mathbf{D^{1}}$ to distinguish it from real data. The whole system is trained adversarially, where the generators and discriminators have opposing aims as far as the value of the objective function is concerned.}
\label{fig:iterative-image-from-text}
\end{figure}

As a simplified example, the vanilla iterative GAN approach could be formulated as a $k$-step generative problem. Without loss of generality, we will consider the case when $k = 2$. The first-stage generator $G^{0}$ and discriminator $D^{0}$ play the already-familiar minimax game:

\begin{equation}
\begin{split}
    \min_{G^{0}}\max_{D^{0}}V(D^{0},G^{0}) = \mathbb{E}_{\mathbf{i}^{0} \sim p_{data}(\mathbf{i}^{0})}\{\log{(D^{0}(\mathbf{i}^{0} \vert \mathbf{t}))}\} \\
    +\mathbb{E}_{\mathbf{z}^{0} \sim p_{\mathbf{z}^{0}}(\mathbf{z}^{0})}\left[\log{(1-D^{0}(G^{0}(\mathbf{z}^{0} \vert \mathbf{t})))}\right]
\end{split}
\end{equation}
where the output of the first generator is $\mathbf{i}^{0} = G^{0}(\mathbf{z}^{0}, \mathbf{t})$. Then, the second stage is the following minimax game:

\begin{equation}
\begin{split}
    \min_{G^{1}}\max_{D^{1}}V(D^{1},G^{1}) = \mathbb{E}_{\mathbf{i}^{1} \sim p_{data}(\mathbf{i}^{1})}\{\log{(D^{1}(\mathbf{i}^{1} \vert \mathbf{t}))}\} \\
    +\mathbb{E}_{\mathbf{z}^{1} \sim p_{\mathbf{z}^{1}}(\mathbf{z}^{1})}\left[\log{(1-D^{1}(G^{1}(\mathbf{z}^{1} \vert \{\mathbf{i}^{0}, \mathbf{t}\})))}\right]
\end{split}
\end{equation}
where $\mathbf{i}^{0}$ and $\mathbf{i}^{1}$ are the images produced in the first and second stage, respectively, $\mathbf{z}^{0}$ and $\mathbf{z}^{1}$ are random variables sampled from pre-determined distributions for each stage, and $\mathbf{t}$ is the text representation. It should be noted that this template is only one of possible setups for the multi-step GAN, as the handling of text representations and the injection of randomness could be potentially done differently, e.g. by varying the text representations between stages and randomizing them, while excluding the latent variable from the second stage. The presented version of the iterative GAN setup emphasizes simplicity and incremental changes from the previously-discussed non-iterative GAN approaches. A simplified layout of the architecture is depicted in Figure \ref{fig:iterative-image-from-text}.

To tackle the problem of generating images from text in a higher resolution than previous models, \cite{Zhang2017stackgan} decompose the generation process into two GAN components in a conditional GAN architecture. Apart from improving the quality of the generated images, the model is designed to handle a specific problem with text representations. Since the text input is mapped to a latent space and various concrete inputs might map to different sub-spaces within this space, it is conceivable that the data manifold for the latent space might exhibit discontinuity, potentially degrading the ability to generate high-quality images. The architecture itself follows the standard GAN template for iterative image from text. Namely, we do have an RNN text encoder and a number of TCNN generators and CNN discriminators. The first GAN component is responsible for generating the rough outline of the final image in lower resolution based on the provided caption. The second GAN is conditioned on the output of the first one and on the caption. It refines the outline to produce the final image with more detail and in higher resolution. To circumvent the problem of the discontinuity in the latent data manifold for text, instead of conditioning directly on the text embedding, the embedding is used to sample latent variables. This is a difference relative to the standard iterative GAN template, as it extends the RNN with the mentioned sampling procedure. This increases the amount of training data and makes the system more robust to small changes in the text data manifold. The method achieves significant improvements over previous approaches, GAWWN \cite{Reed2016draw} and GAN-INT-CLS \cite{Reed2016synthesis}, in terms of quantitative evaluation using the Inception Score \cite{Salimans2016} on the CUB, Oxford-102 and MS COCO datasets. Qualitative evaluations also confirm the ability of the architecture to generate images of higher quality than the previous methods. Additional evaluations show that using two GAN components improves the results relative to using only one component. Similarly, conditioning on sampled text representations also improves the results.

\cite{Bodla2018} propose to approach the task of generating images from text by decomposing the generation process into two stages within a GAN-based framework. The first stage involves two generators. The first one uses random input to generate intermediate representations rather than images. The second generator uses this intermediate representation as input and produces an image.  The second stage introduces a conditional generator. Importantly, the two stages do not necessarily proceed sequentially, but rather the second stage uses a conditional generator, while the two generators from the first stage can be though of as one unconditional generator. Both stages share part of the pipeline for handling random samples from the latent space. This process of reusing part of the information processed by the unconditional generator by the conditional one is a significant difference relative to the standard method. In the second stage, the generator is conditioned not only on the text description but also on the intermediate representation from the first stage. While the general setup of the method follows the standard iterative GAN template, for instance by utilizing an RNN text encoder, there is an important distinction in the use of separate generators for the intermediate representation and the image. An additional difference relative to the standard method is that rather than RNN-encoded text, image attributes can be used as the information to condition on. The architecture compares favorably to StackGAN \cite{Zhang2017stackgan} and other text-to-image architectures in empirical evaluation on the CUB and CelebA \cite{Liu2015} datasets, while requiring less supervisory signal in the data. The results of eyeball tests also suggest that the system is able to generate images of relatively high quality.

A sequential GAN architecture is proposed by \cite{Xu2018}, in which three generators produce progressively more detailed versions of an image. In this way, the model follows the iterative GAN template, also using an RNN text encoder to handle the input descriptions. Each TCNN generator has a corresponding CNN discriminator - yet another similarity to the standard method for iterative GANs. The differences between the proposed setup and the standard approach relate to the handling of the text input by the text encoder, to the processing of features between separate generators, and to an additional alignment between the description and the final generated image. For the text input, not only overall features of the description are obtained, but also features for each specific word. For the processing of features in the pipeline, each layer in the generation process attends to the text input, which allows the system to focus on the most relevant parts of the text at a given generative step. For the alignment between the final image and the input description, a similarity model is proposed to synchronize the image and text features - this is done via two neural networks which learn to align the specific regions in the image with specific parts of the text input. The overall method significantly outperforms previously-available models, including StackGAN++ \cite{Zhang2018} and PPGN \cite{Nguyen2017}, in experiments on the CUB and MS COCO datasets, closing the gap to large scene generation.

A GAN-based text-to-image generative architecture is proposed by \cite{Li2019controllable}. The architecture uses a word-level and channel-wise generator with attention. The intention is to allow the generator to focus on a specific region of the image corresponding to specific words from the description. A word-level discriminator enforces the alignment between the words and image regions. The overall aim is to imbue the generator with the ability to edit regions of the image based on particular words, without affecting the image as a whole. Similarly to the standard iterative GAN template, the proposed architecture utilizes an RNN text encoder to obtain a text feature representation which is then fused with random samples to generate images. The difference is that word features are used via attention in the generation of the image. The image generation pipeline is multilayered with three TCNN generators and three CNN discriminators. Also, a difference relative to the standard iterative method is that each discriminator is not only based on visual features and general text features, but rather incorporates word features as well to determine whether the visual features of the image correspond to the word features of the description. An additional perceptual loss not present in the standard GAN approach is measured based on the final produced image. This loss uses a pre-trained CNN encoder to enforce the alignment of features from real and generated images for a specific layer of the pre-trained CNN encoder. Experiments suggest that the approach is competitive to strong baselines, AttnGAN \cite{Xu2018} and StackGAN++ \cite{Zhang2018}, on the CUB and MS COCO datasets.

\cite{El-Nouby2019} consider the problem of generating the image iteratively, in a series of input steps, where each steps provides a text refining the description of the image. To approach this challenge, they propose a GAN architecture with a bidirectional RNN component to encode descriptions at each time step and an additional RNN component to govern the multi-step generative process. The overall problem and the proposed architecture are considerably different than what is considered in the standard iterative GAN template. First of all, there is an RNN text encoder as is the case for the standard iterative GAN pipeline, but it is used repeatedly at each time step on the incoming refining descriptions of the image. Unlike in the standard method, the second RNN is used to drive the whole process, passing the hidden state to the generator to fuse it with random information sampled from the latent space and produce a new image at each time step. Additionally, a CNN image encoder feeds information about the image from the previous time step to the generator -- an extension to the standard model. Significantly, there are no multiple generators -- the same network is used at each time step to produce the images, which is a significant departure from the vanilla approach. The generated or real image at a given step and the ground truth from the previous time step are processed via yet another CNN image encoder, not present in the standard approach. The fused features from both images are the input to the discriminator, which is also the same for different steps. On top of that, beyond distinguishing between real and generated images, the discriminator has an auxiliary objective of identifying the objects present in the image. Experiments are performed on the CoDraw dataset \cite{Kim2019} and a freshly-constructed dataset named i-CLEVR. An ablation study of the components of the architecture is conducted, revealing that, in general, the components of the model tend to improve the quantitative evaluation metrics. Interestingly, a model without iteration evaluates less favorably compared to the iterative versions. This might suggest that tasks which are not traditionally considered as iterative problems might actually benefit from the iterative approach.

Several problems with the approach to text-to-image synthesis based on the refinement of the initially generated image are highlighted by \cite{Zhu2019}. Namely, the results are sensitive to the initially generated image and, furthermore, text is used indiscriminately while in reality different words have different importance weights attached to them. This is tackled by introducing a GAN architecture linked to a dynamic memory store with read/write gates. The memory storage is supposed to help in situations where the first generated image is hard to decipher and cannot serve as the basis for refinement. The architecture overlaps with the standard iterative GAN template in its use of multiple TCNN generators and CNN discriminators, as well as in the use of an RNN encoder to process the text input. The major difference is the use of the memory structure. The RNN text encoder passes description features to the first generator but it also obtains word-level features and writes them to memory using a gating mechanism. Image features from the previous generative step are also selectively written to memory via this gating mechanism. Each generative step following the initial one is able to read from memory via keys and values using yet another gate. The whole memory-based image refinement part of the pipeline can potentially be repeated multiple times before a final image is produced. The overall use of memory and the details of the approach are significant departures from the standard iterative GAN method. The results from experimental evaluations on the CUB and MS COCO datasets show that for both quantitative and qualitative metrics the approach is able to achieve favorable scores relative to baselines such as StackGAN \cite{Zhang2017stackgan} or AttnGAN \cite{Xu2018}. Sample images from both datasets are presented in Table \ref{tab:cub-mscoco-datasets}.

\begin{table}[t]
  \centering
  \caption{Sample images from the CUB \& MS COCO datasets.}
  \label{tab:cub-mscoco-datasets}
    \begin{tabular}{ccc}
      \toprule
      \multicolumn{3}{c}{CUB \cite{Wah2011}} \\
      \midrule
      \includegraphics[width=2.5cm, height=2.5cm]{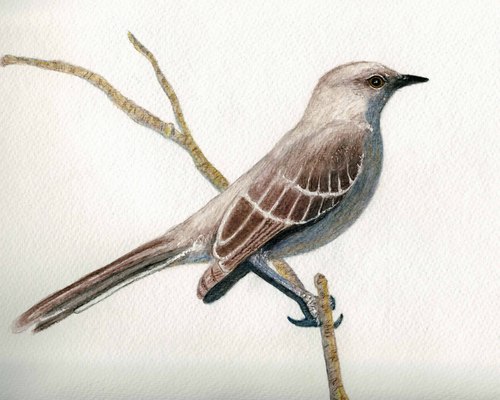} & \includegraphics[width=2.5cm, height=2.5cm]{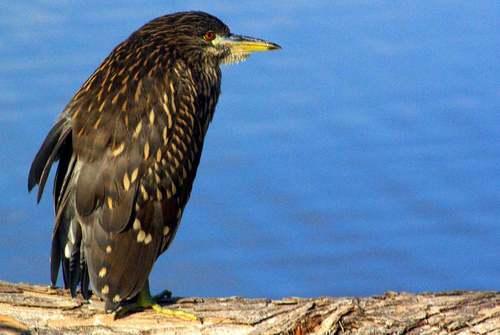} & \includegraphics[width=2.5cm, height=2.5cm]{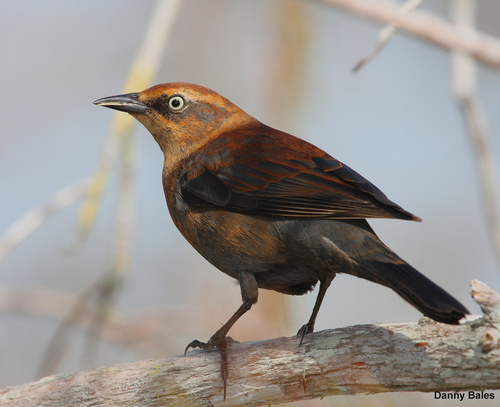} \\
      mockingbird & nighthawk & rusty blackbird \\
      \midrule
      \multicolumn{3}{c}{MS COCO \cite{Lin2014}, \cite{Chen2015}} \\
      \midrule
      \includegraphics[width=2.5cm, height=2.5cm]{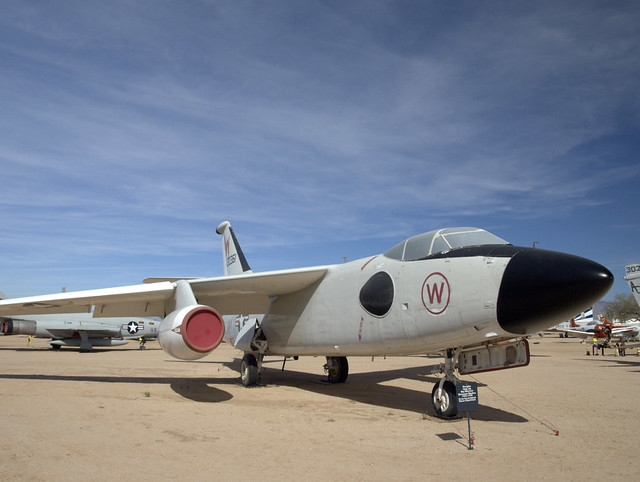} & \includegraphics[width=2.5cm, height=2.5cm]{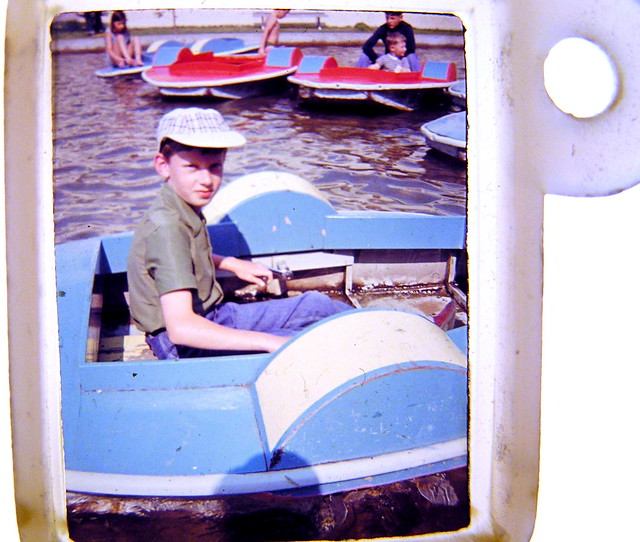} & \includegraphics[width=2.5cm, height=2.5cm]{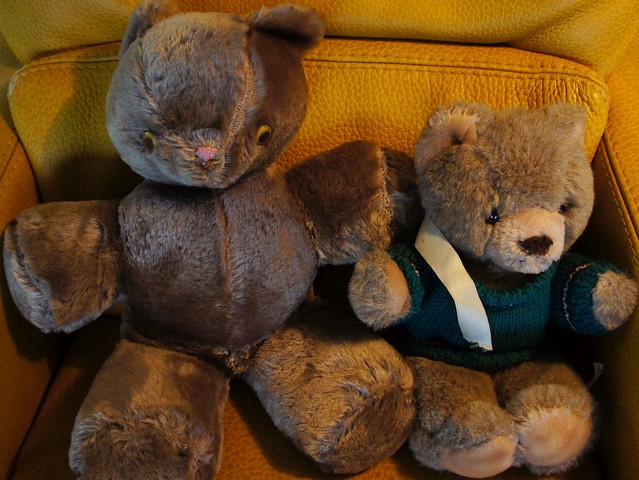} \\
      airplane & person & teddy bear \\
      \bottomrule
    \end{tabular}
\end{table}

Text-to-image generative models frequently employ explicit or implicit text/image alignment procedures. \cite{Qiao2019} attempt to align the images with words in a GAN-inspired architecture. It consists of an encoder, a global/local cascaded image generator for image generation at various levels of detail, and a decoder, which is tasked with regenerating the text description. In line with the standard iterative GAN pipeline, the proposed approach uses an RNN encoder to process the text description to obtain text features. The difference for the text encoder is that it obtains both word-level and sentence level features rather than one representation for the whole description. Apart from that, the word-level and sentence-level features are processed with attention mechanisms to be fused with the randomly generated samples from latent space before being sent to the generator. The process is a multi-stage one, and each stage has its separate generator and discriminator networks. One similarity to the standard iterative GAN approach is the use of TCNN image decoders for each generator and CNN encoders for each discriminator. Unlike in the standard method, an additional CNN image encoder is present to produce the visual features of the generated image and pass them to an additional RNN text decoder which is tasked with the reconstruction of the initial text input based on the visual features of the generated image. Evaluation in an experimental setting on the CUB and MS COCO datasets shows that the proposed model outperforms strong baselines, e.g. PPGN \cite{Nguyen2017} or AttnGAN \cite{Xu2018}. This is true both in quantitative evaluation and in subjective evaluation by human observers.

\cite{Yin2019} focus on the semantics of the input text. They propose a GAN architecture with the specific aim of preserving semantic consistency at a higher level and semantic diversity at a lower level. The high level semantics are learned through a Siamese mechanism \cite{Bromley1993} in the discriminator and the low level semantics are based on batch normalization with conditioning on the input text. The main similarities with the vanilla iterative GAN approach include the use of the RNN text encoder for the processing of the description, the use of a TCNN generator and a CNN discriminator. The image generation is a three-step process with separate generators and discriminators for each stage. In a deviation from the vanilla approach, text inputs are processed in pairs and a contrastive loss is measured for each stage, based on the visual features from each discriminator for the two produced images. Both quantitative and qualitative experimental results on the CUB and MS COCO datasets are suggestive of a distinct advantage of the proposed architecture relative to strong baselines, for example AttnGAN \cite{Xu2018}.

\cite{Liang2020} propose a GAN model to tackle the text-to-image generation problem. They attempt to overcome the problem of translation from the text domain into the visual domain by ensuring that the input text and the synthesized image are consistent on a semantic level. The architecture fits the broad outline of the iterative GAN template, however, it deviates from it in significant ways. One similarity between the methods is the use of a text RNN encoder. A major modification here is that, on top of the text description, the text encoder is made aware of context visual features stored in a memory structure. These context features are obtained based on image detection and attention, where one context feature vector is obtained for one word in the vocabulary. The text input for image generation is processed via an attention mechanism connected to the memory structure. Much like in the standard iterative GAN approach, the generative process follows an upsampling procedure where images of increasing resolution are produced by a set of generators and assessed by a set of discriminators. One difference relative to the standard method is that an additional discriminator is employed for each step, which is tasked with assessing the correspondence of the generated image to the text features obtained from the RNN text encoder. An additional CNN image encoder not present in the standard method is constructed to be aware of the objects present in the image. The features obtained from this image encoder are then used together with the text features from the RNN text encoder to enforce additional alignment between the generated image and the provided text description. Quantitative results on the MS COCO dataset show that the proposed method significantly improves upon very strong baselines, such as DM-GAN \cite{Zhu2019}, AttnGAN \cite{Xu2018} and AttnGAN+ - a custom modification of AttnGAN proposed by the authors.

A GAN-inspired method to enrich the input text description in order to produce images of higher quality is introduced in \cite{Cheng2020}. The overall architecture fits in the standard iterative GAN template, however, the method exploits an attention mechanism to select and match captions from prior knowledge, a mechanism not present in the standard template. A further departure from the template is that rather than overall features from an RNN text encoder, the system uses both description features and word-level features. The enriched captions are used by an attentional GAN which produces an image from a set of captions in a procedure where progressively upsampled images are output by a series of generators and evaluated by a series of discriminators, similarly to the case of the standard pipeline. A significant difference is the use of an attention mechanism to process word-level features from distinct captions. An additional mechanism is also employed to ensure the consistency of the final generated image with the captions. Experiments on the CUB and Oxford-102 datasets show that the model is able to generate images from the enriched captions and that the generated images outperform strong baselines, AttnGAN \cite{Xu2018} and DM-GAN \cite{Zhu2019}, in quantitative evaluations. Qualitative results also suggest that the system is able to generate coherent images of relatively high quality and corresponding to the text input.

The iterative approach to text-to-image generation may also take forms different from the standard GAN- and VAE-based templates. \cite{Wu2022} propose a method which relies on using pretrained image generation models such as StyleGAN2 \cite{Karras2020}. A conditioning module with an energy-based model is used to incorporate additional information in order to arrive at latent codes for the unconditional pretrained model. Text descriptions can be one kind of such additional information. Specifically, the cosine distances between image and text embeddings from the CLIP model \cite{Radford2021} are used as the conditioning information for the text-to-image case. The iterative aspect of the framework comes from the fact that the process can be carried over multiple steps, in which the state of the model at the end of the previous step is used as the initial generative model for the current step, with new conditioning information incorporated at each iteration. This model contrasts with standard iterative text-to-image templates as it does not rely on the explicit passing of image features and whole images between the stages of the process, but rather enables the conditioning information to be changed. Experimental results on the Flick-Faces HQ (FFHQ) dataset \cite{Karras2019} for text-to-image generation show that the proposed architecture produces high quality and diverse images, which compare favorably to the strong baselines of StyleCLIP \cite{Patashnik2021} and StyleGAN2-NADA {\cite{Gal2022}}.

Scaling up and classifier-free guidance \cite{Ho2021} are explored by \cite{Nichol2022}. The overall model follows the standard diffusion template with the major differences being the use of diffusion in pixel space rather than latent diffusion and the addition of a text-conditioned upsampling diffusion model which increases the resolution of the generated images. The model is trained on the same dataset as DALL-E \cite{Ramesh2021}. Evaluations on the MS COCO dataset show that the model outperforms DALL-E \cite{Ramesh2021} and LAFITE \cite{Zhou2022} in terms of quantiative metrics. At the same time, samples generated from the text-conditioned model reflect the meaning of the text and show diversity.

\subsection{Transformer-based image from text}
\label{sec:transformer-image-from-text}

Research on cross-modal text-to-image models has also been influenced by the advances in NLP and vision. One of those advances is the proliferation of attention-based \cite{Bahdanau2015} architectures, in particular, key-value attention models such as the Transformer architecture \cite{Vaswani2017}, first for tasks related to natural language and then, to a lesser extent, for tasks related to vision. 

The Transformer-based methods may follow what could be described as the standard template or the vanilla approach. This would be a setup where we use two separate parts to encode the text and the image. An important distinction here is that both the text encoder and the image encoder are tokenizers, which means that they aim to transform the input text and the processed image into a sequence of tokens rather than a standard vector representation, as is the case in the RNN text encoders and CNN image encoders. For the text tokenizer, approaches such as byte pair encoding (BPE) \cite{Sennrich2016} may be used. Image tokenizers follow a tokenizer-detokenizer setup and may be based on techniques related to discrete VAEs such as VQ-VAE \cite{vandenOord2017} or VQ-VAE-2 \cite{Razavi2019}. In the image tokenizer-detokenizer, a CNN image encoder is the actual image tokenizer and is used to obtain a compressed representation based on tokens. A TCNN decoder is used as the detokenizer to reconstruct the image from the compressed representation. In the standard Transformer-based method, the tokens obtained for both text and image are concatenated and treated as one input stream for an autoregressive decoder Transformer-based network. The Transformer decoder is inspired by autoregressive image models such as Pixel Recurrent Neural Networks (PixelRNN) \cite{vandenOord2016} and their Transformer-based extensions \cite{Child2019}. The delimitation between text and image input is performed via the introduction of specific tokens for marking the beginning of text and image in the sequence. The Transformer decoder is trained to model the joint distribution of text and image tokens. The adoption of such a training procedure enables the architecture to model specific parts of the token input based on other parts of such input. This makes it possible to yield image tokens based on text tokens in a process where the image tokens are autoregressively predicted from the text tokens. Subsequently, the image tokens are used by the image detokenizer, a TCNN image decoder from the image tokenizer-detokenizer model, to produce an image.

In its basic form, the image tokenizer used in the standard Transformer-based template may follow the VQ-VAE architecture \cite{vandenOord2017}, where a discrete latent space $\mathbb{F}$ of features $e \in \mathbb{R}^{K \times D}$ is defined. This discrete latent space may be understood as a \textit{codebook}. $K$ is the size of this space, i.e. the number of feature vectors $\mathbf{e}_{i}$ in $\mathbb{F}$. $D$ is the dimensionality of each feature vector $\mathbf{e}_{i} \in \mathbb{F}$. Like in the VAE architecture, an encoder  $Enc: \mathbb{I} \to \mathbb{R}^{D}$ and a decoder $Dec: \mathbb{R}^{D} \to \mathbb{I} $ are present. $\mathbb{I}$ is the image space. The output of the encoder is denoted by $\mathbf{f}^{Enc} = Enc({\mathbf{i}})$, where $\mathbf{i} \in \mathbb{I}$ is an image and $\mathbf{f}^{Enc}$ is the feature representation of the image obtained from the encoder. In a departure from the standard VAE model, the distribution $q( z = k \vert \mathbf{i})$ is defined as:

\begin{equation}
    q(z = k \vert \mathbf{i}) =
    \begin{cases}
        1, & \text{for}\ k = \argmin_{j}{{\Vert \mathbf{f}^{Enc} - \mathbf{e}_{j} \Vert}_{2} }\\
        0, & \text{otherwise}
    \end{cases}
\end{equation}

The representation $\mathbf{f}^{Enc}$ is then replaced by the nearest element from the discrete latent space:

\begin{equation}
\label{eq:vq-vae}
    \mathbf{f} = \mathbf{e}_{k},\ \text{where}\ k = \argmin_{j}{{\Vert \mathbf{f}^{Enc} - \mathbf{e}_{j} \Vert}_{2} }
\end{equation}

The final reconstruction of the image is produced by passing the discretized features to the decoder $\mathbf{i}^{rec} = Dec(\mathbf{f})$. Since Equation (\ref{eq:vq-vae}) is not differentiable, a straight-through gradient estimator \cite{Bengio2013} can be applied. A constant KL divergence term is obtained by assuming a uniform distribution over $z$. The final training objective is as follows:

\begin{equation}
\label{eq:loss}
    \mathcal{L} = \log{p_{\theta}(\mathbf{i} \vert \mathbf{f})} + {\Vert \text{sg}[ \mathbf{f}^{Enc} - \mathbf{e} ] \Vert}_{2}^{2} + \beta {\Vert \mathbf{f}^{Enc} - \text{sg}[\mathbf{e}] \Vert}_{2}^{2}
\end{equation}
where $\mathbf{sg}$ is the stop gradient operator. The first term of Equation (\ref{eq:loss}) is the reconstruction loss. The second one is a Vector Quantization term which aligns the discrete latent space with the output of the encoder. The third term is a commitment loss, which constraints the ability of the encoder to grow its output and ensures that the encoder commits to a representation.

In the overall standard Transformer-based template, the image encoder from VQ-VAE is used to tokenize the images. The final image output in the template is produced autoregressively and may follow the PixelRNN architecture \cite{vandenOord2016}. The main idea is to model a probability distribution of image pixels:

\begin{equation}
    p(\mathbf{i}) = \prod_{i=1}^{n^{2}}{p(x_{i} \vert \{ x_{1}, \dots, x_{i-1} \})}
\end{equation}
where $\mathbf{x}$ is the overall $n \times n$-pixel image and $x_{i}$ is the $i$th pixel in the image. Similarly to language models, the next prediction is based on the previously-predicted elements. The actual pixel predictions are modeled using CNN and LSTM \cite{Hochreiter1997} layers, which, depending on the concrete type of the LSTM layer, process input in rows or diagonally to output predictions for specific pixels.

The decoder is based on the Transformer architecture, which employs multi-layer and multi-head attention mechanisms. Attention is defined as:

\begin{equation}
    A(\mathbf{Q}, \mathbf{K}, \mathbf{V}) = softmax\left( \frac{\mathbf{Q}\mathbf{K}^{T}}{\sqrt{d_{k}}} \right)\mathbf{V} 
\end{equation}
where $\mathbf{Q}$, $\mathbf{K}$, $\mathbf{V}$ are all matrices. $\mathbf{Q}$ can be understood as queries, $\mathbf{K}$ as keys used for comparisons with queries and $\mathbf{V}$ as values which are combined linearly to produce the final output. The component $d_{k}$ denotes the dimensionality of the keys and is employed as a safeguard against the dot product $\mathbf{Q}\mathbf{K}^{T}$ growing large in magnitude and pushing the softmax function into regions where its gradients are small.

In a Transformer, attention is applied independently multiple times in parallel, yielding multi-head attention:

\begin{equation}
    MHA(\mathbf{Q}, \mathbf{K}, \mathbf{V}) = concat(\mathbf{head}_{1}, \dots, \mathbf{head}_{h})\mathbf{W}^{O}     
\end{equation}
with $\mathbf{head}_{i} = A(\mathbf{Q}\mathbf{W}_{i}^{Q}, \mathbf{K}\mathbf{W}_{i}^{K}, \mathbf{V}\mathbf{W}_{i}^{V})$. $\mathbf{W}_{i}^{Q}$, $\mathbf{W}_{i}^{K}$ and $\mathbf{W}_{i}^{V}$ are learnable projection matrices and $h$ is the number of attention heads to apply in parallel. The Transformer architecture employs several other mechanisms for handling input data and combining multi-head attention in an encoder-decoder structure \cite{Vaswani2017}.

\begin{figure}[h]
  \centering
  \includegraphics[width=0.7\linewidth]{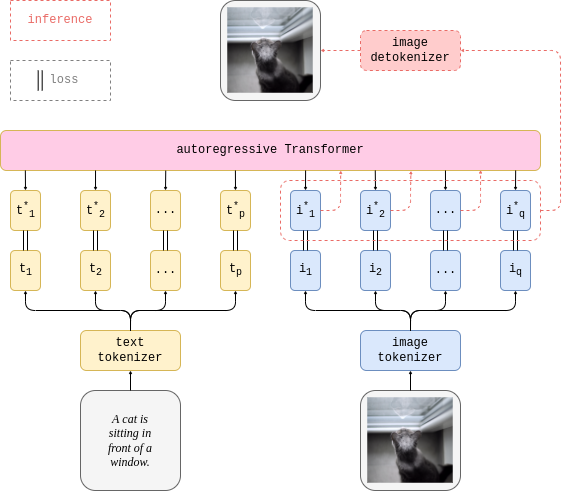}
  \caption{Standard Transformer-based image from text template. Separately to the main training loop, a generative image model comprised of an image tokenizer and detokenizer is pre-trained to compress an image into a series of tokens (image tokenizer) and reconstruct the image from the tokens (detokenizer). Similarly, a text tokenizer is trained to produce a sequence of text tokens based on text input. In the main training procedure, a series of text tokens is produced from text input by the text tokenizer and a series of image tokens is obtained from image input by the image tokenizer. An autoregressive Transformed model learns to model the text and image tokens as a single stream of data. At inference, no image input is required as all the image tokens are predicted. The predicted image tokens are passed to the pre-trained image detokenizer to generate the final image.}
\label{fig:transformer-image-from-text}
\end{figure}

A schematic overview of the Transformer-based text-to-image template is presented in Figure \ref{fig:transformer-image-from-text}.

A comparison of the evaluation results of Transformer-based text-to-image models on the MS COCO dataset is presented in Table \ref{tab:mscoco-transformer}. The relative sparsity of the Transformer-based research field and the lack of unified standards for the evaluation of such models limit the number of architectures for which comparable results are reported.

Building upon Transformer-based advancements, a zero-shot text-to-image generative framework is proposed by \cite{Ramesh2021}, also known as DALL-E. The proposed architecture relatively closely follows the Transformer-based standard template. In particular, they use a setup related to discrete VAEs such as VQ-VAE \cite{vandenOord2017} or VQ-VAE-2 \cite{Razavi2019}. An encoder is trained to compress an image into a grid of tokes, which are then concatenated with encoded text tokens and used to train an autoregressive Transformer network - all in line with the standard template. In quantitative experiments, the model is trained on a newly-constructed dataset of 250 million text-image pairs, and then evaluated on the MS COCO and CUB datasets against the following baselines: AttnGAN \cite{Xu2018}, DM-GAN \cite{Zhu2019} and DF-GAN \cite{Tao2022}. For MS COCO, the system is able to compete with the baselines and outperform them in the zero-shot evaluation setting. The quantitative results on the CUB dataset are far worse, with the approach significantly underperforming, perhaps due to the highly specific nature of the data distribution for CUB. In human evaluations on MS COCO, in upward of 90\% of majority votes the system is assessed as producing images better matching the caption or being more realistic than those produced by the methods chosen for comparison.

The adoption of Transformers for text-to-image generation is shown in \cite{Ding2021} to be successful in terms of the improvement in the quality of the produced data. This work is another example of an architecture closely adhering to the standard Transformer-based template. It has much overlap with \cite{Ramesh2021}, while being developed independently. The approach adopts an autoregressive Transformer network which uses both text tokenization and image tokenization techniques. In particular, the image tokenizer is based on VQ-VAE, while the text tokenizer is based on SentencePiece \cite{Kudo2018} tokenizer. Both text and image tokens are concatenated such that both the text and image tokens are treated as parts of the same sequence in a pretraining procedure which requires the model to predict masked parts of the mentioned sequence. All of this conforms to the standard template. There are some minor differences from \cite{Ramesh2021}, such as the weighting of text vs. image tokens in the loss for training, which is done equally rather than weighting image tokens more heavily. Similarly, training stabilizing techniques are applied, which make it possible to train the model with mostly half precision (FP16) rather than full precision (FP32). The model can also be finetuned on several tasks such as super-resolution, style learning, etc. Quantitative results on MS COCO show that the model outperforms strong baselines, including \cite{Ramesh2021}, by a large margin on a number of metrics and, in particular, is significantly more favorably scored by human observers than GAN-based text-to-image baselines.

\begin{table}
  \caption{Text-to-image Transformer-based generation, MS COCO. The presented evaluation metrics are described in Section \ref{sec:text-to-image}.}
  \label{tab:mscoco-transformer}
  \resizebox{0.7\textwidth}{!}{
    \begin{tabular}{ccccccccc}
      \toprule
      Method & IS & FID-0 & FID-1 & FID-2 & FID-4 & FID-8 & CapLoss & comments\\
      \midrule
      \cite{Ramesh2021} & 17.9 & 27.5 & 28.0 & 45.5 & 83.5 & 85.0 & - & as reported by \cite{Ding2021} \\
      \cite{Ding2021} & 18.2 & 27.1 & 19.4 & 13.9 & 19.4 & 23.6 & 2.43 &  - \\
      \bottomrule
    \end{tabular}
  }
\end{table}

In a Transformed-based architecture, \cite{Zhang2021ufc} propose to adopt non-autoregressive generation. They also consider the ability of the model to not only include text or visual information but also constraints requiring the produced image to preserve specific parts of the input image. In broad strokes, the approach follows the standard Transformer-based template, with the significant deviation being the replacement of the Transformer autoregressive decoder with a non-autoregressive decoder. Model training can be divided into two stages as in the standard pipeline. During the first stage, the model utilizes tokenizers for text and image processing. A VQ-GAN-inspired tokenizer is used for the image part. BPE \cite{Sennrich2016} is applied to obtain text tokens. In the second stage, the tokens are formed into a sequence and the system is trained to predict the masked tokens, which follows the standard approach. There is a minor difference relative to the standard pipeline in the fact that two additional tasks are used in the training procedure: binary classification of the relevance of the generated image and a similar binary classification task for the fidelity of the generated image. The major difference relative to the vanilla method comes from the images being generated in a non-autoregressive manner based on token masking and using the relevance and fidelity scores obtained in the two binary classification tasks. The generation comes in a series of masking and token prediction steps, with the process being stopped based on the relevance and fidelity scores. A new dataset, M2C-Fashion is constructed. The proposed architecture is evaluated on this dataset and the Multi-Modal CelebA-HQ dataset \cite{Xia2021}. A quantitative comparison of results to a number of strong baselines, including VQGAN \cite{Esser2021taming}, shows that the proposed method beats the baselines, including the autoregressive VQGAN approach, for almost all considered metrics. The lack of an autoregressive component helps the model achieve greater generation speed than its autoregressive counterparts. It also allows the model to incorporate bidirectional information and constraints on the specific parts of the output.

\subsection{Self-supervised}
\label{sec:self-supervised}

Advances in self-supervised learning \cite{Zbontar2021}, \cite{Bardes2022} have made it an attractive area for research related to sample efficiency due to the potential of self-supervised methods to significantly limit the amount of labeled data needed for training. The ability to achieve results comparable to fully-supervised methods with a fraction of the labels has fueled the investigation of such architectures for vision. The applications in cross-modal settings are few and far between but nonetheless first efforts have been made. So far, those methods have mostly relied on contrastive \cite{Chopra2005} approaches.

Since the amount of work on text-to-image self-supervised learning is limited in comparison to other branches of text-to-image generation, there is no standard template which can be used to compare the specific methods. The closest we are able to come to such a template is the observation that the superivised approaches are usually linked to VAEs or GANs. As such, they reuse some familiar components. For VAE-based methods, such components would include a CNN image encoder, an RNN text encoder, a TCNN image decoder and an RNN text decoder. For GAN-based methods, a TCNN decoder/generator and CNN encoder/discriminator are present along with an RNN text encoder. These methods incorporate the self-supervised approach via techniques related to contrastive learning. One example of such a technique would be obtaining representations from the text and image modality for both real and generated samples and performing comparisons between samples corresponding to real or generated text-image pairs, as well as similar samples which do not come from the same pair. Quite notably, VAE-based text-to-image self-supervised methods might rely on more general architectures which do not necessarily assume the specific modalities a priori, such as the MVAE \cite{Wu2018mvae} and MMVAE \cite{Shi2019}. The particulars of specific architectures vary.

The training of models based on multiple modalities usually requires considerable amounts of paired multimodal data. To circumvent the need for extensive paired datasets, \cite{Shi2021relating} proposed a VAE-based method with contrastive elements. This means that the model is able to learn not only from the explicit links between modalities but also by the distinction between related and unrelated data from different modalities. The contrastive aspect is expressed via a triplet loss where the difference between the mutual information calculated for related and unrelated examples from two modalities is maximized. The MNIST-SVHN dataset is constructed based on the MNIST and SVHN \cite{Netzer2011} datasets. This new dataset, along with the CUB dataset is used for the evaluation. The proposed method is found to improve the process of multimodal training, in particular by limiting the necessary amount of data relative to JMVAE \cite{Suzuki2017}, MVAE \cite{Wu2018mvae} and MMVAE \cite{Shi2019}.

A weakness of self-supervised generative models is highlighted by \cite{Sutter2021} in the form of a trade-off between learning the joint data distribution and semantic coherence. Two dominant approaches in the multimodal generative space: MVAE \cite{Wu2018mvae} and MMVAE \cite{Shi2019} are shown to be different only in the specifics of the joint posterior approximation functions. The authors design a more general architecture, of which the two mentioned are special cases. This architecture is aiming to allow favorable modeling of the joint distribution (MVAE) and semantic coherence (MMVAE) at the same time. The generalized framework is then shown to compare favorably to MVAE and MMVAE in terms of quantitative evaluations for the text-to-image tasks on the newly-constructed PolyMNIST dataset, as well as on the MNIST-SVHN-Text and CelebA datasets (both introduced in \cite{Sutter2020}).

\cite{Zhang2021contrastive} design a text-to-image GAN contrastive network. The generator produces an image from a caption. At the same time, the information from a real image paired with this caption is used for guiding the learning process. A generator, image encoder and text encoder are trained together in a procedure where both the generated and real images are processed by the image encoder and the caption is processed by the text encoder. The main idea is to guide the training such that the embeddings from the image encoder for the real and generated image are in agreement with one another and also with the embeddings for the caption produced by the text encoder. Conversely, once a different image/caption pair is considered, the embeddigns for this image are once again steered to remain close but the embeddings between the two image/caption pairs are guided to be distinct. The Localized Narratives \cite{Pont-Tuset2020} and MS COCO datasets are used to construct the LN COCO dataset. The Localized Narratives and the OpenImages \cite{Kuznetsova2020} datasets are further used to obtain the LN OpenImages dataset. On the MS COCO, LN COCO and LN OpenImages datsets, the model shows quantitative results competitive with strong baselines including, among others, AttnGAN \cite{Xu2018}, DM-GAN \cite{Zhu2019}, SD-GAN \cite{Yin2019} and CPGAN \cite{Liang2020}. Human evaluators also reveal to have a strong preference for images generated by this method compared with strong baselines. Qualitatively, the images produced by the method display relatively higher quality than those of other analyzed architectures.

\subsection{Video from text}
\label{sec:video-from-text}

A quite natural extension of text-to-image generation is the process of generating videos from a text description. In principle, this can be seen as a series of text-to-image steps with additional complexity stemming from the fact that not only each of those images individually have to meet the requirements of the text-to-image task but the generated video as a whole should also correspond to the text description while retaining coherence and diversity. All this makes for an even more challenging problem setup.

The standard template for generating video from text shares a lot of components with the standard approach to generating images from text. In particular, we can distinguish between approaches based on VAEs and those based on GANs.

The VAE text-to-video standard approach includes an RNN text encoder to process the input text and obtain features which are then used to arrive at the parameters needed to sample from the latent space. The sampled representation is then processed by an RNN to produce representations for multiple time steps, which correspond to the frames of the produced video. The representation at each step is then fed into a TCNN image decoder which outputs the particular frame. The RNN together with the TCNN form the overall RNN-TCNN video decoder, which produces frames in sequence to finally obtain the whole output video. The produced video can then be measured against the actual video from the video-text pair from the dataset via a reconstruction loss. As in the case of the standard VAE, a KL divergence term ensures that the distribution of the latent variable is consistent with a presupposed distribution, which enables sampling at inference.

The GAN text-to-video standard approach also includes an RNN text encoder, which obtains the text features and those features are used in the process of sampling. The random representation, together with the text features is then processed by an RNN. Similarly to the VAE case, the RNN produces representations for each time step, which are fed into a TCNN image decoder to arrive at the image for a given time step. These two components form an RNN-TCNN video decoder/generator, which is, again, in line with the VAE approach. A CNN image encoder is used to process each frame of a real or generated video and the representation at each time step is processed by an RNN to finally arrive at a score which determines whether the particular video is real or generated. The CNN image encoder and the RNN together form a CNN-RNN video encoder/discriminator. This discriminator also has access to the input text feature representation. The system is trained adversarially, where the RNN-TCNN generator competes with the CNN-RNN discriminator.

In both the GAN and the VAE templates, an RNN encoder is used to obtain the representation of the input text $f$. The RNN text encoder processes the text input $\mathbf{t}$, to obtain a representation of the text in the \textit{visual-semantic space} $\mathbf{f}$:

\begin{equation}
\label{eq:text-rnn-encoder}
    \mathbf{h}_{t}^{Enc} = f^{Enc}(\mathbf{h}_{t-1}^{Enc}, \mathbf{t}_{t})
\end{equation}
where $f^{Enc}$ is the part of the overall encoder responsible for the hidden state and $\mathbf{t}_{t}$ is the word at time step $t$ in the encoding phase. The output of the encoder is $g^{Enc}$:

\begin{equation}
    \mathbf{f} = g^{Enc}(\mathbf{h}_{t_{end}}^{Enc})
\end{equation}
where $t_{end}$ is the last time step of the encoding phase. In this formulation, the whole input text is summarized as a fixed-length representation $\mathbf{f}$.

Both the GAN and VAE templates also share the structure of the decoder/generator in the vanilla case. Let us denote by $f^{Dec}$ the part of the RNN decoder responsible for modeling the hidden state $\mathbf{h}_{t}$ of the decoder at time step $t$. The hidden state is updated as follows:

\begin{equation}
    \mathbf{h}_{t}^{Dec} = f^{Dec}(\mathbf{h}_{t-1}^{Dec}, \mathbf{f})
\end{equation}

At each time step, the features $\mathbf{v}_{t}$ of a frame of the output video are generated:

\begin{equation}
    \mathbf{v}_{t} = g^{Dec}(\mathbf{v}_{t-1}, \mathbf{h}_{t}^{Dec}, \mathbf{f})
\end{equation}
where $\mathbf{v}_{t-1}$ are the features of the previous generated frame of the video. The features of each frame are passed to a TCNN decoder for the generation of the actual frame. A special token is used to signal the end of the video.

\begin{figure}[h]
  \centering
  \includegraphics[width=0.65\linewidth]{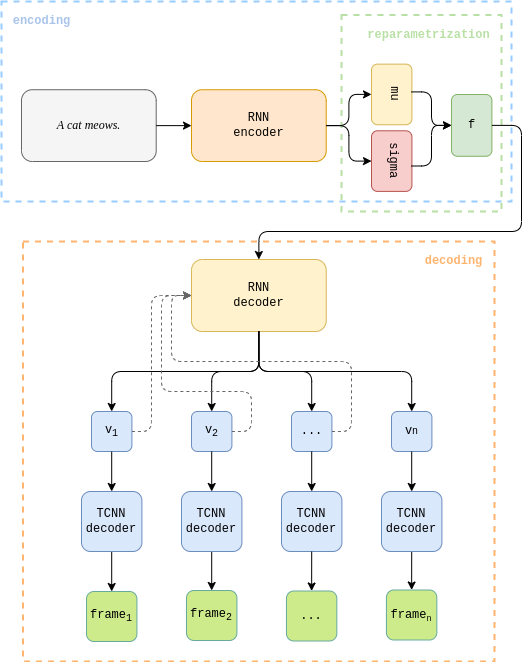}
  \caption{Standard video from text VAE template. Text input is processed by an RNN decoder to obtain parameters $\mu$ and $\sigma$ which are used to sample a feature representation $\mathbf{f}$. This representation is fed into an RNN decoder which produces video features $\mathbf{v_{t}}$ for each time step $t$ based on the representation and the video features from the previous time step. The video features are passed to a TCNN decoder which generates a video frame for the specific time step.}
\label{fig:video-from-text-vae}
\end{figure}

An outline of the standard video from text VAE-based template is shown in Figure \ref{fig:video-from-text-vae}. Alternative schemes, where the text encoder outputs features $\mathbf{f}_{t}$ at each time step are also possible, but we omit them for brevity.

It has to be noted that, due to the complexity of video data, specific approaches might significantly deviate from the standard templates as they might employ custom mechanisms to enforce the consistency or diversity of the generated video.

As far as diffusion-based models are concerned, text-to-video generation is a very active research area and the approaches within it build on various incarnations of diffusion architectures. Consequently, there is no single standard diffusion template for text-to-video generation.

In an attempt to generate videos from captions, \cite{Marwah2017} propose a conditional VAE architecture in which captions are used together with short- and long-term visual dependencies between video frames to generate videos of variable length. The overall architecture follows the standard VAE template, with several deviations. An RNN text encoder is present as in the standard pipeline. One significant difference is that the architecture employs separate attention mechanisms linking the captions to the short- and long-term dependencies. The short-term dependencies are obtained via a CNN image encoder based on the previous generated video frame. Long-term dependencies are obtained via a CNN image encoder applied to all the previously-generated frames, followed by an RNN. These two form a CNN-RNN video encoder, which obtains the representation for the video generated so far. Both short- and long-term representations are combined with the text representation from the RNN text encoder and processed via the mentioned attention mechanisms to be passed to the RNN-TCNN video decoder at each time step to generate the frame of the video at this step. Two datasets are used for evalutation: a specifically-constructed version of Moving MNIST \cite{Srivastava2015}, i.e. Two-Digit Moving MNIST, and the KTH Human Action Database \cite{Schuldt2004}. The results show that the proposed system is able to generate videos from previously-unseen captions and that the generated videos at least partially match the caption. It is also shown that conditioning on both short- and long-term context is beneficial as far as quantitative evaluation is concerned. The system is able to generate videos of variable length. Additionally, switching the caption mid-generation changes the characteristics of the generated video while at least partially preserving the coherence of the generated frames.

\cite{Balaji2019} approach the same general problem of text-to-video generation. A GAN-based architecture is proposed in which a series of images is generated sequentially by the generator through an RNN. The proposed setup shares many components with the standard GAN approach with several differences. Among the similarities, it uses an RNN text encoder and an RNN-TCNN image decoder/generator. The first important difference relative to the vanilla approach is that a random vector is sampled from the latent space at each time step rather than at the beginning of the video generation. Another significant modification relative to the standard approach is that two discriminator networks are employed, rather than the standard one. The first one assesses individual video frames, the second one attempts to assess the whole video. Yet another modification comes in the form of the inclusion of two image encoders/discriminators instead of one. One of these discriminators operates on individual frames, while the other one on the video as a whole. Each of these networks is divided into a series of sub-networks, CNNs or not, where these networks process the image sequentially. The output of each of these sub-networks is passed to the next sub-network and also to an additional network linked to this specific sub-network. The aim of the linked networks is to produce the output of the sub-part, which is then concatenated with the outputs of the other sub-parts of the discriminator to form a representation, which is in turn used to determine whether the presented frame/video is real or fake. The overall model is evaluated quantitatively on the Moving Shapes dataset, created specifically for this evaluation, and on the Kinetics dataset \cite{Li2018video}. The results show that it improves upon previous architectures, such as T2V \cite{Li2018video}, as far as evaluation metrics are concerned. They also suggest that the improvement relative to simplified GAN approaches, e.g. conditional GANs with text-video feature concatenation, is substantial.

\cite{Deng2019} base their solution on a GAN architecture, where the generator utilizes an RNN-TCNN component, in line with the prescriptions of the standard GAN approach. A major difference relative to the standard approach is that the RNN part of the generator does not feed into the TCNN through one representation, but rather passes information to each TCNN layer. This enforces the generator to emphasize the details of each individual frame while taking into account the temporal coherence of the video as well. Similarly to the standard approach, the architecture uses an RNN text encoder. A CNN discrimiator is used as in the standard approach but its details vary considerably. It encompasses two sub-networks, where one of them is a 3D CNN network, which operates on the level of the whole video. The other one is a CNN which operates on frames and specifically incorporates the difference between the representation of the current and the previous frame. Both the video and frame sub-networks also reconstruct the text representation from the RNN text encoder. In a further change to the standard technique, a mutual information approach is utilized in order to align the generated video with the provided description. This is done via the alignment of the reconstructed text representations with the representation from the RNN text encoder. Three datasets are constructed for evaluation: Single Moving MNIST-4, Double Moving MNIST-4 and KTH-4 (the last one reconstructed based on the method described in \cite{Mittal2017}). Experimental evaluation shows that the proposed architecture compares favorably to strong baselines, including Sync-DRAW \cite{Mittal2017} and MoCoGAN \cite{Tulyakov2018}, in terms of quantitative evaluations.

\cite{Hayes2022} introduce MUGEN - a large-scale video-audio-text dataset, which can be potentially used in various kinds of cross-modal or multi-modal generation tasks. In particular, it is shown that it is feasible to train and evaluate video-from-text architectures on the proposed dataset. A model related to the standard VAE template for video-from-text is trained to generate videos from text. The model differs from the vanilla template in the use of an autoregressive Transformer for the output part. Qantitative results for variations of the training setup suggest that the dataset is in fact suitable for the evaluation of video-from-text generative models.

An architecture incorporating quantization is proposed by \cite{Han2022}. It deviates from the standard video generation templates. While the text RNN encoder is still relatively consistent with the vanilla methods, the model uses VQ-VAE \cite{vandenOord2017} to obtain quantized visual features in place of a standard VAE encoder. Additionally, the text and visual features are processed by a Transformer network. The representations obtained from the Transformer network can be decoded by the VQ-VAE decoder used to obtain initial visual representations. Overall, the system has common elements with the VAE template, but the modifications are relatively pronounced. The whole architecture is trained on three tasks: masked sequence modelling, relevance estimation and video consistency estimation. Evaluations of the designed method are performed on the Moving Shapes dataset \cite{Balaji2019}, the MUG dataset \cite{Aifanti2010}, the iPER dataset \cite{Liu2019liquid} and the new Multimodal VoxCeleb dataset constructed from the VoXCeleb \cite{Nagrani2020} dataset. Quantitative evaluations show that the model is competitive with strong baselines such as TFGAN \cite{Balaji2019} on the considered datasets.

A diffusion-based approach is put forward in \cite{Ho2022}. This architecture largely follows the standard template for diffusion text-to-image generation, with two significant differences. First, it adjusts the encoder-decoder architecture in the reverse diffusion process in order to operate on a sequence of frames of predetermined length rather than on a single image. Second, it does not run the diffusion process in the latent space but rather applies the forward process directly in pixel space, similar to earlier diffusion works \cite{Sohl-Dickstein2015}, \cite{Ho2020}. The video generation process is conditioned on BERT \cite{Devlin2019} embeddings. A number of modifications to the training process are considered, such as: joint training on image data, classifier-free guidance \cite{Ho2021} and an autoregressive extension. The training is performed on an internal dataset of 10 million captioned videos. The results suggest that the considered extensions to the core process are beneficial from the point of view of quantitative evaluation metrics and in terms of the quality of the generated samples.

\subsection{Image editing}
\label{sec:image-editing}

The process of generating images from text does not necessarily have to be restricted to producing an image based on a caption alone. The image editing or image manipulation task can be thought of as a text-to-image problem where the input consists not only of the text description but also of the image to be manipulated. Inherently, this might be an easier problem than generating images from text alone, as the input image contains significant information about the desired output. At the same time, image manipulation systems face the challenge of disentangling the features and relating concepts from the text to specific regions of the image. It has to be noted that there is also a significant difference on the text side. While for previous text-to-image methods the text provided the description of the generated image, here it may provide such a description or it might provide the description of the edits that are to be applied to the image. 

The fact that the task setup for image editing is relatively similar to the one for text-to-image generation, at least on the basic level, makes the problem amenable to approaches described for text-to-image generation. And so, VAE-based, GAN-based and diffusion-based approaches are popular also for image editing tasks.

\begin{figure}[h]
  \centering
  \includegraphics[width=0.8\linewidth]{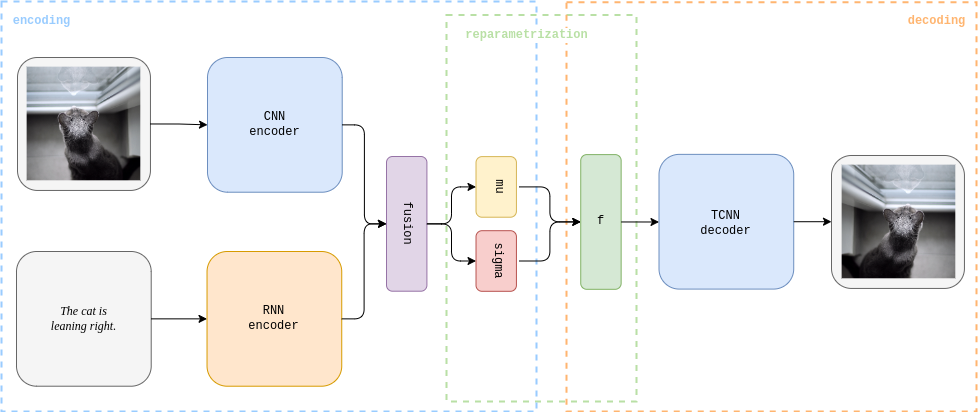}
  \caption{Standard image editing VAE template. An input image, which is the basis for the edit, is processed by a CNN encoder to obtain visual features. Similarly, text input with the information about the edit is handled by an RNN encoder to produce text features. The text and visual features are fused into one feature representation, and this representation is used to obtain parameters $\mu$ and $\sigma$. These parameters are used to sample a representation $\mathbf{f}$, which is passed to a TCNN decoder to produce the final edited image.}
\label{fig:image-editing-vae}
\end{figure}

The VAE template includes an RNN text encoder to obtain a feature representation for the input text. A CNN image encoder is used to process the input image and get the visual features. The text and visual features from the text and image encoders are then used to obtain the parameters used in the sampling of latent representations. These representations are then fed into a TCNN image decoder, which is used to produce the edited image. As is the case for plain text-to-image, a reconstruction loss is imposed on the actual and reconstructed image and the training procedure enforces the adherence of the parameters obtained from the encoder to a predetermined distribution via measures such as the application of KL divergence. An outline of such an architecture is presented in Figure \ref{fig:image-editing-vae}.

The GAN template for image editing is relatively similar to the template for text-to-image generation. Once again, we have an RNN text encoder, which is used to arrive at text features which are then used along with randomly sampled representations to form a representation fed into a TCNN image decoder/generator. A separate CNN image encoder/discriminator is tasked with processing the real and produced images and determining whether a particular image is real of generated. It takes the image in question as input along with the text features obtained from the text RNN encoder. The system is trained adversarially, with the generator and discriminator competing. As is the case for VAE-based approaches and for image editing more generally, the major difference here is the inclusion of the image and its visual features in the encoding process.

One of the effects of the rapid progress in diffusion-based image editing methods is the lack of a standard template for diffusion image editing. Particular methods present significantly divergent approaches to the task, with one specific difference being the process formulation via: an inpainting mask and text, text-only description, text-only edit instruction. Consequently, the descriptions of the specific diffusion image editing techniques do not refer to a standard diffusion image editing template but may compare their components to the vanilla text-to-image template.

\cite{Nam2018} address the discussed problem of modifying an image based on the provided text description. They design a GAN-based architecture, which overall follows the standard GAN approach for image editing, but in which the discriminator is modified to assess whether the information related to a specific word in the caption is present in the processed image. This information is used to produce the final score of the discriminator. The architecture uses an RNN text encoder and a CNN image encoder, as is the case in the standard template. A difference here is that before feeding the joint text-visual representation to the TCNN decoder, the representation is processed by a series of residual blocks with a single skip connection. The mentioned modifications to the discriminator are significant and are the distinguishing feature of this approach relative to the standard one. First, the discriminator uses a separate RNN text encoder to process the text input and does not reuse the RNN text encoder linked to the generator. Second, the discriminator is decomposed into a series of sub-networks which are to determine whether there is visual correspondence between the image and a given word of the input text. The final score from the overall discriminator is obtained via an attention mechanism. The architecture is evaluated on the CUB and Oxford-102 datasets and compared with two baselines: SISGAN \cite{Dong2017} and AttnGAN \cite{Xu2018}. Human evaluations reveal that the system is favorably scored by users. Qualitative evaluations suggest that it is able to modify images in line with the meaning of the caption while preserving the context not related to the caption, such as the background of the image.

\cite{Li2020lightweight} design a lightweight GAN-based architecture with a word-level discriminator. In the broad outline, it follows the standard approach. There is an RNN text encoder, which is used to produce the representation of the provided description. All the other usual components are present, albeit with significantly modified details. Instead of one CNN image encoder, there are two separate ones, with differing CNN architectures. The visual features from the image encoders are fused with text features in a pipeline consisting of upsampling and residual blocks, text-image affine combination modules and attention, rather than a standard TCNN from the vanilla GAN approach. The disciminator network is also modified. It takes an input image and text features as in the standard pipeline, but it relates the input adjectives and nouns to the regions of the image and obtains word-weighted image features. Such a discriminator provides the generator with a learning signal at the word level, which allows the generator to focus on the parts of the image which are related to the attributes described in the text, without modifying the rest of the image. Experiments on the CUB and MS COCO datasets show that the lightweight nature of the architecture allows it to retain competitive levels of performance relative to a strong baseline, ManiGAN \cite{Li2020manigan}, while also significantly limiting the number of trained parameters.

The specific form in which the text input is used as an editing request coupled with an image is tackled by \cite{Jiang2021}. They propose a GAN-based method for editing the global characteristics of an image based on a text request. The main difference relative to the other approaches described so far is that the architecture operates on editing requests rather than image descriptions. This means that the input text describes the edit rather than the desired output - a setup admitted by the standard GAN template. The overall proposed architecture differes significantly from the standard GAN approach. As far as similarities are concerned, an RNN encoder is used to process the text edit request and obtain a representation for it. Similar to the standard approach, this representation is passed to a generator along with the original image to produce an edited version of the image. Other than in the standard GAN method, the generator has an embedded attention mechanism and the edited image is processed together with the original one via a separate CNN-based network to produce the edit embedding. A cycle consistency loss and reconstruction loss are applied in a cyclic framework - not present in the standard GAN pipeline. On top of that, the input and target images might be swapped or randomly augmented. A similarity to the standard GAN method comes in the form of a CNN discriminator used to score the images. The GIER \cite{Shi2020} and MA5k-Req \cite{Shi2021learning} datasets are used for experiments. Quantitatively, the system achieves scores higher than those of strong baselines, e.g. PixAug \cite{Hai2018} and OMN \cite{Shi2020}. Qualitative results also support the system's ability to perform the editing of an image as far as global image characteristics are concerned.

General text-to-image generation and image editing do not necessarily have to be handled by distinct architectures. Text-based image generation and text-based image editing are unified in one framework in \cite{Xia2021}. The proposed architecture is significantly different from the standard GAN template in multiple aspects. An RNN text encoder is used but the embeddings obtained from it are not independently fed into a generator but rather learned jointly with the embeddings of the CNN image encoder to map into a shared text-visual space. This means that the visual and text embeddings for an image with its description are brought close together. Other than in the standard GAN approach, the generator itself is based on the StyleGAN \cite{Karras2019} generator rather than the usual TCNN. A further departure from the standard method is the presence of a separate network which is tasked to invert a pretrained GAN generator to encode an image in the latent space of the generator. Image editing involves obtaining both the text request's and the manipulated image's embeddings in the joint latent space through image and text encoders. These two embeddings are mixed to obtain a new embedding for the image after manipulation. Not present in the standard GAN approach, additional instance-level optimizations are performed to obtain the embedding used to generate the image. Yet another salient departure from the vanilla GAN method is that a control mechanism with attribute-specific selection is used for controlling the flow of information into the generator and unifying the text-to-image and image editing tasks in one common architecture. A new dataset is introduced for evaluation: the Multi-Modal CelebA-HQ dataset. For image generation, AttnGAN \cite{Xu2018}, ControlGAN \cite{Li2019controllable}, DM-GAN \cite{Zhu2019} and DF-GAN \cite{Tao2022} are considered for comparison. For image editing, ManiGAN \cite{Li2020manigan} is used as a baseline. As far as the results of the experiments are concerned, the proposed architecture is able to quantitatively outperform all the analyzed baselines on both the image generation and image editing tasks. Overall, the results show that the images generated from this method are of relatively high quality for both text-to-image generation and text-based image manipulation. The generation itself is also shown to potentially produce images of relatively high diversity.

One potential extension to the image editing task is to not limit the model to generating a new image based on the reference one and an input text, but to allow the architecture to perform the mirror operation of generating new text based on the reference one and an input image. In practice, for an input image and text such a system would output a new image, guided by the input text, and a new text, guided by the input image. This is the overall idea of the architecture proposed by \cite{Li2022}. It is based on the GAN template for image editing with adjustments made for the fact that a text is generated as well – e.g. an additional text discriminator. Additionally, the method relies on the construction of graphs for both the input image and text. These graphs are used as input for the semantic alignment of representations. Such semantically-aligned representations are fed into a graph masking mechanism for relation reasoning. Experimental results on the CUB \cite{Wah2011} and Visual Genome \cite{Krishna2017visual} datasets show that the method outperforms strong baselines such as SISGAN \cite{Dong2017}, AttnGAN \cite{Xu2018}, TaGAN \cite{Nam2018}, ManiGAN\cite{Li2020manigan}, while also generating images and text which correspond to the provided text/image input.

In traditional text-to-image and image editing methods there is no explicit structure for influencing the semantically meaningful regions of the generated image. \cite{Shi2022} attempt to imbue a GAN architecture with such capabilities. They broadly follow the GAN template for image editing, however, with significant differences. In particular, rather than obtaining one random representation to be processed by a GAN generator, the architecture maps multiple random representations into a common space and processes each of those representations by a separate generator. Each generator produces both a feature map and a pseudo-depth map and the maps from all the generators are fused into a general segmentation mask and a general feature map. A generator network is fed the feature map and generates both a refined version of the segmentation mask and the actual image. A discriminator network is used much like in the standard template, however, it operates on both the images and their segmentation maps. Qualitative results of text-guided image generation on the CelebAMask-HQ dataset \cite{Lee2020} suggest that the proposed method is able to generate diverse and consistent images while applying edits to semantically meaningful parts of the image, even when the images are edited sequentially.

The general text-to-image diffusion model presented in \cite{Nichol2022} can be adapted to perform image editing tasks by allowing the architecture to take inpainitng masks as input. A model trained on the dataset from \cite{Ramesh2021} is shown to be able to perform qualitatively appealing and consistent image edits, which can be chained in sequences of edit actions.

The conditioning on text input in diffusion image editing models does not necessarily have to take place via injecting the conditioning information into the reverse diffusion model. \cite{Kim2022diffusion} follow a research line where a forward diffusion model is used in pixel space to obtain a noise representation of the input image, while a reverse diffusion model is used to recover an image, albeit not the original one but rather one that corresponds to a supplied text prompt. Image and text representations are obtained via CLIP \cite{Radford2021}. This departs from the standard text-to-image diffusion template, where a latent diffusion model is used and the inverse diffusion model is directly conditioned on the text representation. Training and evaluation is performed on the CelebA-HQ \cite{Xia2021}, AFHQ-Dog \cite{Choi2018}, LSUN-Bedroom and LSUN-Church \cite{Yu2016} datasets. Qualitative evaluations against TediGAN \cite{Xia2021}, StyleCLIP \cite{Patashnik2021} and StyleGAN-NADA \cite{Gal2022} suggest that the model is capable of generating relatively appealing and consistent imagery. It is also preferred by human evaluators to StyleCLIP and StyleGAN-NADA and outperforms these two methods in quantitative trials.

In localized editing of an image based on text input, it is only a specified part of the image that is to change, while the majority of the pixels should remain unchanged. \cite{Avrahami2022} propose a method that displays these precise characteristics. The overall approach is based on masking the edited part of the image and using it as input to the diffusion process, together with a representation of the edit prompt. Different than in the standard text-to-image diffusion approach, it is the pixel space that the diffusion process operates in. On top of that, the diffusion process itself is modified to include a separate forward diffusion path, where the original non-masked image is made progressively noisier. The images from this separate path are blended with the main diffusion process for the masked image. The method uses CLIP \cite{Radford2021} as a pre-trained text encoder. The diffusion model itself is trained on ImageNet. The performance of the proposed system is assessed against PaintByWord \cite{Bau2021} and a VQGAN \cite{Esser2021taming} method with CLIP and PaintByWord. Visual examination of the produced image edits for a given mask and text prompt suggests that the method is able to generate more appealing and consistent samples than either of the approaches it is compared with. Also, a user study shows that human evaluators score this method higher than the baselines on overall realism, background preservation and the correspondence between the prompt and the generated image.

Diffusion-based image editing techniques frequently require the user to provide an input map in order to determine which parts of the image are to be edited and which are to be left untouched. \cite{Hertz2023} provide a method to perform localized edits based on text only, without the need for input maps. The architecture itself follows the standard text-to-image diffusion template, but it operates on pixels rather than in latent space. Also, rather than one diffusion process, it employs two: one for generating an image from the original text prompt and one for generating an edited image from an edited text prompt. In addition, cross-attention is used to relate the text input with specific pixels in the image. Edits are made by fusing the attention maps from the original and the edited diffusion process. Three edit functions are considered for this, based on the kind of edit that the user wants to make: a word swap edit, a prompt refinement edit and an emphasis edit. The system also has a process to obtain an edit mask without user input when localized edits are specifically requested. The method, in its main version, uses a pre-trained Imagen \cite{Saharia2022} model and does not require additional model re-training. Evaluation data is produced based on pre-determined text edit templates and the used pre-trained model. Comparisons are run against: the VQGAN \cite{Esser2021taming} with CLIP, Text2Live \cite{BarTal2022} - both text-guided, as well as Blended Diffusion \cite{Avrahami2022} and GLIDE \cite{Nichol2022} - both inpainting-mask-guided. Qualitative evaluations suggest that the proposed method is able to edit images based on text edits alone, and that the results are visually appealing and more consistent than in the compared approaches. A user study is run to assess the generated samples on three axes: (1) background and structure preservation, (2) alignment with text, and (3) realism. The study compares the proposed method against VQGAN with CLIP and Text2Live and concludes that users prefer the proposed system to the other methods for all three aspects.

Within the diffusion-based image editing realm, \cite{Brooks2023} take an approach bootstrapping from the results of prior models. They use the Prompt-to-Prompt model \cite{Hertz2023} and GPT-3 \cite{Brown2020} to generate a dataset of over 450,000 training examples, where each example consists of the original image, an image modified by Prompt-to-Prompt, and the corresponding text edit prompt. The unique characteristic of this dataset is that the conditioning text is now an edit instruction, not a description of the desired edited image. An architecture largely adhering to the text-to-image diffusion template is proposed, with Stable Diffusion being the core model \cite{Rombach2022}, and a modification in the form of the inclusion of not only text but also image conditioning information in the reverse diffusion process. This extension is necessary to handle the setup where an image is provided together with an edit text. Apart from that, the method includes classifier-free guidance \cite{Ho2021}. The results of training the model are qualitatively compared with several baselines: SDEdit \cite{Meng2022}, Text2Live \cite{BarTal2022} and Prompt-to-Prompt \cite{Hertz2023}. These comparisons suggest that the proposed method has significant advantages over the other ones regarding its ability to generate images consistent with the edit instructions and consistent in their internal structure. Apart from that, it is the only method which is able to operate directly on edit prompts. Quantitative comparisons with SDEdit and Prompt-to-Prompt on metrics measuring image consistency and edit quality show that for the same level of image consistency, the proposed method achieves higher levels of edit quality than the other approaches. In particular, it outperforms Prompt-to-Prompt, which has been used in its training procedure.

The possibilities of text-only image editing models are investigated further by \cite{Mokady2023}. In particular, their focus is on applying editing models to real-world images rather than to ones generated by a diffusion model. To this end they propose an inversion procedure where the forward diffusion process is reversed by DDIM inversion \cite{Song2021}, \cite{Dhariwal2021}, however, this DDIM inversion is treated as a pivot \cite{Roich2021} around which the actual optimization procedure operates. In particular, the reverse process utilizes classifier-free guidance \cite{Ho2021} where the unconditional null text embedding is optimized and the final predicted latent code for a given step is pushed toward the pivot, i.e. the representation from the DDIM inversion process. Even though the approach significantly modifies the text-to-image diffusion template, the core of the vanilla method is still present in the form of a Stable Diffusion model \cite{Rombach2022} which is used as the backbone of the proposed method. The actual editing process uses Prompt-to-Prompt, without retraining this model. Experimental results on a subset of the MS COCO evaluation set show that the model is able to successfully reconstruct images beating a number of baselines, such as the DDIM inversion process itself and the encoder-decoder part of Stable Diffusion. For the image editing task itself, qualitative assessment on images used by \cite{BarTal2022} shows that the method generates appealing and consistent imagery relative to other text-editing approaches: VQGAN \cite{Esser2021taming} with CLIP \cite{Radford2021}, SDEdit \cite{Meng2022} and Text2Live \cite{BarTal2022}. A user study finds the proposed method outperforming these approaches in terms of fidelity to both the original image and the text edit. An additional user study on ImageNet suggests that the proposed model outperforms SDEdit, Text2Live and FlexIT \cite{Couairon2022} in terms of the accuracy of edits, but not necessarily in terms of similarity to the original image. Qualitative tests also show similar results for a comparison with inpainting-mask-based methods: Stable Diffusion \cite{Rombach2022} itself, Blended Diffusion \cite{Avrahami2022} and GLIDE \cite{Nichol2022}.

A slightly different take on diffusion-based image editing formulates it as a task of editing real images but without the need for a priori specification of editing types and without the need for synthetically generated images. At its core, the method uses the vanilla text-to-image diffusion template, albeit it does not specifically require it to be a latent diffusion model. Apart from that, the optimization and inference procedures introduce significant changes. Two specific stages are introduced for the traing procedure. In the first one, the diffusion model is treated as pre-trained and not optimized. An input image is treated with the diffusion model together with the desired text description of the image after the edit. This is in contrast to methods such as \cite{Brooks2023} where the text is the edit. The text embeddings are optimized to allow for the reconstruction of the input image. In the second stage, the text embeddings are frozen, and the pre-trained diffusion model is fine-tuned with the same reconstruction objective. Inference is performed based on the input random noise as in standard diffusion approaches, however, it also includes a linear interpolation between the initial unoptimized text embedding and the final optimized text embedding. Two specific models are used as the core diffusion components in experiments: Stable Diffusion \cite{Rombach2022} and Imagen \cite{Saharia2022}. The method is applied on a variety of images collected by the authors from Unsplash and Pixabay - two stock photography services. Importantly, the whole training procedure involves orders of magnitude less computing resources than other text-to-image methods. Qualitative results show that the trained architecture is able to perform a variety of involved edits, while preserving the internal consistency of the images. Comparisons with Text2Live \cite{BarTal2022}, SDEdit \cite{Meng2022} and DDIB \cite{Su2023} point to the possibility that the designed model preserves more consistency in the image than the other methods, while successfully applying the edit. The Text Editing Benchmark (TEdBench) dataset is introduced for evaluation purposes. A user study shows that a significant majority of users prefer images generated by the proposed approach to images generated by any of the compared methods.

\subsection{Graphs}
\label{sec:graphs}

Text-to-image generation methods tend to rely on generating the whole image from the provided text or on manipulating the image based on the text. An extension in which the input text is not presented in a raw form but rather structured in a graph representation can also be considered. In this problem setup, additional information about the relationships between various parts of the text input can be utilized to obtain more relevant representations and images of higher quality.

Since graph approaches to text-to-image generation are an active area of research, there is yet to emerge one template method with which all the ongoing work could be contrasted. It should be noted, however, that there are specific components, which emerge within the graph-to-image line of research. Most notably, the input to the model is a graph, which necessitates the use of graph convolutional networks \cite{Duvenaud2015}, \cite{Kipf2017} to obtain a graph representation. In current methods, this representation is then used to obtain the crops of objects represented in the graph structure. These crops are processed by a CNN image encoder and fed into a TCNN image decoder. Depending on the specific approach, the whole system may be VAE- or GAN-based. 

\cite{Li2019pastegan} construct a GAN-inspired architecture which is able to generate images from scene graphs in order to enable more control in the generation process, such as explicit manipulation of specific objects in the image and not the image as a whole. Based on a scene graph and object crops delivered as input, the model is able to generate whole images incorporating the information from both the graph and the crops. Experiments suggest that in quantitative evaluations this method is able to perform considerably better than very strong baselines: sg2im \cite{Johnson2018} and layout2im \cite{Zhao2019}.

A GAN-based architecture to parse a textual layout and generate an image with a scene corresponding to this layout is proposed in \cite{Ashual2019} and \cite{Ashual2020}. In these works, the scene graph is transformed into embeddings of each object in the graph, bounding boxes for each object are created and appearance information is added. Based on this information, the final image is generated. In qualitative evaluations, the technique demonstrates the ability to parse and successfully generate relatively coherent images for progressively complicated layouts. The quantitative results show that the architecture has the potential to improve upon the results of sg2im \cite{Johnson2018} and layout2im \cite{Zhao2019} - previous baselines.

\subsection{Other}
\label{sec:other}

While significant swaths of the text-to-image research area fall into one of the above-mentioned sub-fields, the presented categories are by no means exhaustive. There are other lines of research which do not neatly fit into any one of these categories. Nonetheless, these research directions constitute interesting avenues, and so, even a glimpse at the possible other approaches to text-to-image generation could be worthwhile, even if, by definition, there are no common standard templates for those methods. 

An interesting sub-field within text-to-image generation is the task of generating realistic handwriting from input text. \cite{Kang2020} approach the problem of generating handwriting images from text and images of handwriting. In a GAN-based architecture, three learning objectives are introduced: to produce realistic images, to imitate handwriting styles presented in the input images and to be consistent with the provided text input. The generator is conditioned on the text input and the images of written words. The produced handwriting is then passed to a discriminator, to a classifier for writer recognition and to an encoder-decoder RNN for word recognition. Experiments show that the proposed architecture is able to generate handwriting which is indistinguishable for humans from real handwriting.

\cite{Li2021} address the problem of generating talking heads tailored to a specific user, based on text input, by designing a two-stage GAN-inspired architecture. In the first stage, time-aligned text is used to produce parameters relating to the head pose, upper face and mouth shape characteristics. This speaker-independent information is then fed into the second-stage speaker-specific model. This model incorporates an attention network to generate face landmarks based on the received characteristics. The obtained landmarks are then used by yet another network to generate videos. Quantitative results for the sub-modules relating to the upper head and to the mouth suggest that these components achieve favorable scores on evaluation metrics relative to strong baselines, for instance, pix2pixHD \cite{Wang2018synthesis} and CSG \cite{Sadoughi2021}. Qualitative results relative to approaches including, i.a., Wav2Lip \cite{Prajwal2020} and NVP \cite{Thies2020} also suggest that the proposed method is able to generate talking head of relatively high fidelity. Additionally, in human evaluations, the videos generated from this system are evaluated as real in more than 50\% of the cases, which outperforms all the analyzed baselines, relative to over 90\% for real videos.

Rather than simply generate frames from text, \cite{Liu2022beat} have tackled the related task of generating human gestures from multi-modal input data. Their architecture has similarities with the GAN video-from-text template, but it accepts multiple inputs: text, speaker ID, emotion labels, audio data and images. Each handled input modality has its specialized encoder. On the text input side, the architecture uses a temporal convolutional encoder rather than a classic RNN one. On the output side, rather than using the RNN-TCNN decoder, the model applies separate body and hand decoders and one discriminator. The system is trained adversarially, with additional loss components for the reconstruction of body and hand gestures. A new dataset is specifically constructed for the experiments – the Body-Expression-Audio-Text (BEAT) dataset. The proposed network outperforms a series of strong baselines such as MultiContext \cite{Yoon2020} in quantitative comparisons.

\begin{figure}[h]
  \centering
  \includegraphics[width=0.8\linewidth]{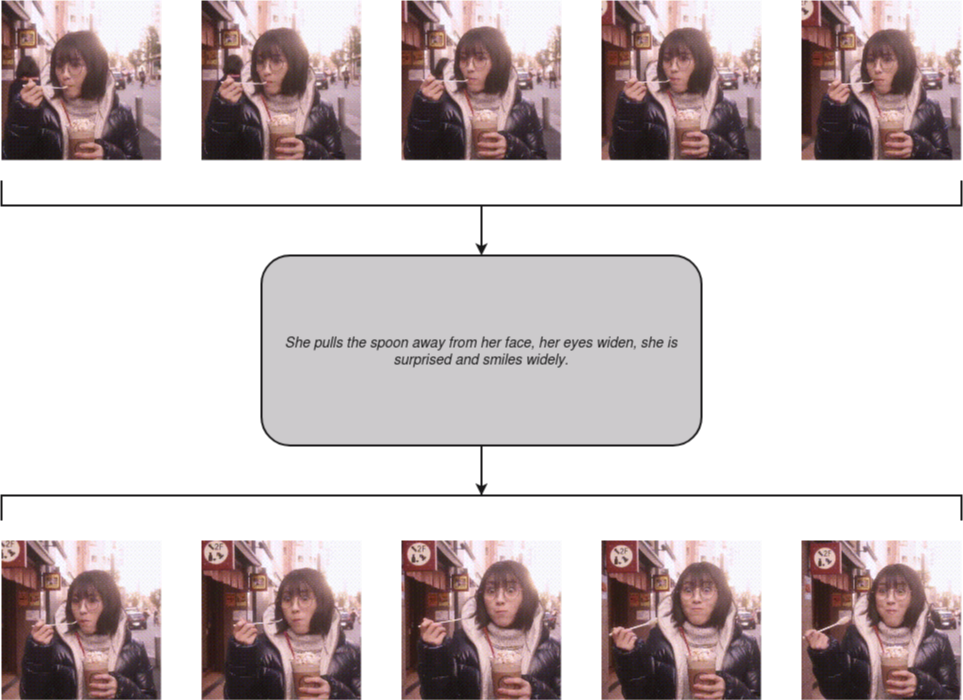}
  \caption{Overview of the video editing task. Video frames come from the FERV39k dataset \cite{Wang2022}.}
\label{fig:video-editing}
\end{figure}

While generating video from text and image editing have been popular research lines, the related task of video editing is only starting to enjoy increased attention. \cite{Fu2022} tackle precisely the problem of editing videos based on text input. An overview of this task is presented in Figure \ref{fig:video-editing}. The proposed approach shares similarities with both the video-from-text GAN template and the image editing GAN template. The text input is processed by an RNN encoder as in the former and the individual input video frames are fed into a CNN encoder as in the latter. There are significant departures from the two standard templates, though. The RNN encoder models both the word- and sentence-level representations. Those representations are handled by a Transformer network together with the video frame features and spatial coordinates. This network produces features to be decoded by a frame generator network, which is similar to the generator in the standard templates. The generated video frames are submitted to two discriminator networks in a further departure from  the standard methods. One of those discriminators operates on the frame level and the other one on the level of sequences of consecutive frames. Three new datasets are built for evaluation purposes: E-MNIST, E-CLEVR and E-JESTER. Results on these datasets show that, in quantitative terms, the proposed architecture performs favorably to strong baselines such as E3D-LSTM \cite{Wang2019e3d}.

\section{Directions for future research}
\label{sec:future}

\textit{Cross-modal generation} is at a point where it has seen a significant increase in the interest of the research community, while at the same time a number of its sub-fields remain relatively weakly covered, particularly when compared with the most popular one - \textit{image captioning}. Such a combination of factors makes cross-modal generation a potentially fruitful area for future research. In particular, there is significant potential in the less developed areas which are rapidly getting more attention. This includes text-to-image.

Paired datasets remain the dominant for most cross-modal generation problems. For tasks linking the text and visual domains, they usually come in the form of images or videos paired with text descriptions. Both text-to-image and image-to-text methods rely on such data. The problem with using such datasets is that there is an inherent limit to how much this approach can scale. In particular, the construction of large, high-quality datasets of images with their descriptions requires some form of manual annotation. The problem is more significant than that, though. For problems of high dimensionality, e.g. understood as the size of the output space, it is not feasible to construct large enough datasets in any reasonable time frame. For instance, the output space for the problem of generating images from text encompasses multiple possible values for each considered pixel and there are potentially infinitely many valid pictures for a specific input text. The lack of sufficiently large aligned datasets puts a natural cap on classic approaches. One remedy might be the development of methods which do not explicitly rely on paired data. Examples of this include models based on domain adaptation \cite{Chen2017} or weakly-supervised learning \cite{Laina2019}. Such approaches may help alleviate the problem, however, significantly more work needs to be done in order to bring the results of methods such as weakly-supervised ones in line with what is currently possible for fully-supervised models. Also, the research into such methods tends to focus on the image-to-text area, which makes it even more important to further work on such methods for text-to-image as well.

The introduction of additional structure to the latent space of text/visual architectures has been one of the most fruitful research directions in terms of the improvement of evaluation metrics. This trend has been visible for models working specifically with the text/visual data, but also for more general architectures, in particular, ones which do not necessarily assume specific modalities, but rather are constructed in ways which allow the modalities to be swapped without modifying the core architecture. The latter general approaches proposed in the field of cross-modal generation tend to rely on latent space decomposition or on a specific learning objective \cite{Shi2019}, \cite{Sutter2020}, \cite{Mahajan2020}. These approaches show some promise as far as the quality of the generated samples is concerned. This notwithstanding, the design of general cross-modal architectures is still very much an open problem. One possible way forward is to consider further approaches to latent space decomposition. This in itself is a challenging problem, as any such design cannot rely on the structure of any specific modalities, as the general nature of the architecture requires it to be possible to change the modalities without modifying the architecture. The methods relying on general latent space decomposition tend to be VAE-based. This opens yet another possibility - one of proposing GAN- or diffusion-based architectures geared toward the same general task. This could potentially result in an improvement in the quality of the generated data, but requires significant research into training techniques.

An interesting focal point is the possibility of using text data from different languages. Multilingual cross-modal architectures have enjoyed some interest \cite{Wang2021efficient}, with attempts to utilize unpaired samples \cite{Chen2019bilingual}. Given the relatively limited nature of hand-made datasets and the problems with scaling supervised approaches on such data, there is significant potential in the area focusing on the move from classic paired supervised training examples to more loosely coupled data points. The multilingual problem is a natural way to increase the amount of available data. For instance, for the problem of generating images from a text description, the inclusion of descriptions in languages other than the target one could significantly increase the size of the available datasets. The main problem here is the design of architectures able to handle different languages. The extent to which such an architecture would have to change between languages, and how much of it could actually be general, are interesting research questions.

There has been some effort to incorporate knowledge broader than what can be learned from the training dataset itself. One particular area of application is visual dialogue, where external knowledge bases have been proposed \cite{Yang2021unimf}, showing that it is possible to rely on such sources of information. This has, however, been limited to specific cases. An attempt to apply similar techniques for text-to-image generation is one potential avenue of research, with the particular challenge being that while in VQA (Visual Question Answering) the external knowledge might come in the form of text, for text-to-image other knowledge sources, in particular visual ones, could be required. The application of text-to-image methods need not be limited to VQA. Other areas of Visual Reasoning might also benefit from the adoption of approaches generating images from text, a prominent example being the field of Abstract Visual Reasoning \cite{Malkinski2022raven}, \cite{MalkinskiMandziuk2023}, where the ability to generate visual data might find its use, for instance, in completing ambiguous sequences.

Text-to-image generation does not have to be the end task. Cross-modal generation admits various kinds of inter-connectivity between its sub-areas. For instance, it is possible to train a text-to-image model to solve an upstream task and use the generated data for a downstream VQA task to improve results of a VQA model \cite{Yang2021dialogue}, as VQA systems are still relatively far from human-level performance \cite{Talmor2021}. The challenge for this specific setup relates to the fact that the images obtained from a text-to-image architecture are still likely to lack the quality and diversity of the samples from the actual image dataset. There is a gap as far as research on improving quality and diversity of images for downstream tasks is concerned. Text-to-image models could also be on the receiving end of data augmentation procedures. One example could be a system for the handling of text data, which is trained to replace parts of a description with their synonyms. Such a text model could be used as a preprocessing pipeline for a text-to-image model.

A prominent trend is the idea of iterative improvements to the generated images \cite{Zhang2017stackgan}, \cite{Bodla2018}, \cite{Xu2018}, \cite{El-Nouby2019}, \cite{Zhu2019}, \cite{Liang2020}, \cite{Cheng2020}. In such methods, hierarchical output generation pipelines are used. It would potentially be of interest to utilize a hierarchical input representation as well and verify whether enforcing input representations at different levels of the hierarchy to correspond to the improvements introduced by the subsequent decoders could be beneficial for the system. As an example, it would be interesting to see whether it is possible to construct a system in which the first level of input representation is related to specific parts of the text and higher levels are more abstract, and make these layers of abstraction inform the generative part of the system, where the first generated image is conditioned on the abstract representation and the subsequent ones introduce more specific input representations.

One of the most important developments in text-to-image generation in recent years is the rapid shift from VAEs and GANs to diffusion. The initial introduction of diffusion for image generation \cite{Sohl-Dickstein2015}, along with improvements to the techniques \cite{Ho2020}, \cite{Dhariwal2021} and the advent of latent diffusion architectures \cite{Rombach2022} has seen a tectonic shift in terms of a transition towards diffusion. The field of text-to-image generation is no different with significant progress being made. There is still a lot of potential in exploring the scaling properties of diffusion models and, in particular, in investigating the potential to combine existing models in zero-shot settings.

The propagation of diffusion has significantly impacted image editing research where the standard stable diffusion model \cite{Rombach2022} has been shown to be easily adaptable to image-editing tasks. Consequently, diffusion methods in various shapes and forms \cite{Hertz2023}, \cite{Brooks2023}, \cite{Mokady2023}, \cite{Kawar2023} have become the most prominent line of investigation. The main aspects that would benefit from being addressed are the consistency of the generated images and the broadening of the range of potential edits.

Text-to-video generation has seen significant improvements \cite{Marwah2017}, \cite{Balaji2019}, \cite{Deng2019}, although one specific aspect of video generation seems to be underrepresented in this research area. Namely, given a text description and the video stream up to a given moment, it is possible to generate multiple plausible completions of the video. Further research in the text-to-video space could explicitly try to model such a range of possible outcomes. It would also be interesting to see methods to construct datasets useful for such architectures explicitly handling ambiguity in the output. Exploring the applicability of diffusion models to this domain seems like a particularly promising endeavor.

Another avenue of research within the text-to-video domain could involve text-based video editing. Such methods have already been shown effective for text-to-image \cite{Nam2018}, \cite{Li2020lightweight}, \cite{Jiang2021}, \cite{Xia2021}. An extension of this research to handle video streams would involve changing the input video based on the provided description. Going from images to videos would require the systems to not only be able to generate individual modified frames, but also to carry over the information from the input description and apply it in a temporal manner where not all the required modifications/instructions would be applied at each time step. This could constitute a fruitful line of investigation. The same goes for the incorporation of diffusion methods, especially image-editing models. The enforcement of temporal consistency for diffusion might be a potential highlight.

The limited amount of labeled data and the need to extend the learning algorithms to data without explicit labels has led to a significant amount of work being done on self-supervised methods \cite{Zolfaghari2021}, \cite{Shi2021relating}, \cite{Zhang2021contrastive}, largely in the contrastive learning paradigm \cite{Chopra2005}. These methods have shown the applicability of self-supervised approach in cross-modal learning, but they suffer from the drawbacks of contrastive approaches, namely the need for hard negative samples, which limits the scalability of such methods. Because of that, it might be of interest to investigate non-contrastive self-supervised approaches, which have already shown some degree of applicability to single modalities \cite{Zbontar2021}, \cite{Bardes2022}. To the best of our knowledge, there is no direct research on text-to-image non-contrastive self-supervised methods. In our opinion, this constitutes a research opportunity with much potential.

It should be noted that text-to-image generation, in contrast to image-to-text, faces the significant challenge of modeling images, which currently requires the modeling of raw pixel values on the output end. A lot of the research effort relies on including additional structure in the generative process and there is still a lot of room for the continuation of this kind of work. Significant research opportunities might also lie in the addition of structure on the output end of the generative pipeline, namely for the specific step of image production. Rather than rely on pixels themselves, intermediate representations tied to graph structures or semantic meaning could be used before offloading the final work of producing the actual pixels. The extent to which this is a viable option for image generation remains an area of potential research.

\section{Conclusions}
\label{sec:conclusions}

In this work, we have reviewed the state of research in the cross-modal text-to-image generation domain. To this end, we have analyzed papers published at 8 leading machine learning conferences in the 2016-2022 time frame, along with additional relevant papers. We have found that the field is still underdeveloped in terms of publication output relative to the more popular image-to-text area, yet it has seen both a significant increase in interest and considerable progress. This combination of factors makes it a particularly interesting area of future research.

We have proposed to partition the field of text-to-image generation into several relevant sub-fields. We have discussed: the vanilla text-to-image area, its iterative and Transformer-based extensions, self-supervised methods, the text-to-video area, image-editing, graph methods, and the remaining idiosyncratic approaches. The iterative and Transformer-based methods are the core of the current advances in text-to-image cross-modal generation. Actual generative models in the like of VAEs, GANs and diffusion are dominant within all the analyzed fields. In particular, diffusion methods have recently seen a substantial increase in the quality of produced samples and in the interest of the research community.

Text-to-image generation remains less covered than image-to-text methods, potentially due to the fact that the output space is significantly more involved than for image-to-text approaches, as the space of potential valid images given a description is, in principle, significantly larger than the space of valid descriptions given an image. This complexity of the output space also results in the application of generative models to this problem, as non-generative approaches are relatively limited in their ability to produce convincing visual data. Frequently, these methods are based on diffusion, GANs or VAEs to produce the actual visual data. Research generally focuses on topics such as generating images or video from text, with specific problems including iterative and non-iterative generation, Transformer-based models, self-supervised methods, image edition and graph inclusion. The fact that text-to-image generation has received increased attention and shown successful approaches, but the number of published papers is still relatively limited but increasing suggests that this is an interesting area of research.

The text-to-image domain has shown significant promise, yet considerable research gaps have been identified as well. It is possible to pursue incremental lines of research, as well as to investigate the possibility of designing novel architectures, the applicability of self-supervised methods, zero-shot approaches and other avenues of inquiry. We are of the opinion that text-to-image cross-modal generation may receive further elevated levels of interest, similarly to what has been the case for image-to-text methods. Work on general architectures and data efficiency improvements might take center stage.

The factors considered in this review point to the text-to-image area as a relatively young one, with increasing numbers of successful approaches, many existing research gaps, significant practical impact and a potential to bring together methods designed for single-modality settings. With interesting lines of research, and the potential for new ones to crop up, text-to-image cross modal generation is one of the most interesting areas in contemporary machine learning research.


\bibliographystyle{ACM-Reference-Format}
\bibliography{text-to-image}
\end{document}